\documentclass[fleqn,10pt]{wlscirep}
\usepackage[utf8]{inputenc}
\usepackage[T1]{fontenc}
\usepackage{bbm}
\usepackage{bm}
\DeclareMathAlphabet{\mathcal}{OMS}{cmsy}{m}{n}
\pdfoutput=1
\title{Uncertainty-Aware Mixed-Variable Machine Learning for Materials Design}

\author[1]{Hengrui Zhang}
\author[1]{Wei (Wayne) Chen}
\author[1]{Akshay Iyer}
\author[2]{Daniel W. Apley}
\author[1,*]{Wei Chen}
\affil[1]{Department of Mechanical Engineering, Northwestern University, Evanston, IL 60208, USA}
\affil[2]{Department of Industrial Engineering \& Management Sciences, Northwestern University, Evanston, IL 60208, USA}

\affil[*]{email: weichen@northwestern.edu}

\begin{abstract}
Data-driven design shows the promise of accelerating materials discovery but is challenging due to the prohibitive cost of searching the vast design space of chemistry, structure, and synthesis methods.
Bayesian Optimization (BO) employs uncertainty-aware machine learning models to select promising designs to evaluate, hence reducing the cost.
However, BO with mixed numerical and categorical variables, which is of particular interest in materials design, has not been well studied.
In this work, we survey frequentist and Bayesian approaches to uncertainty quantification of machine learning with mixed variables.
We then conduct a systematic comparative study of their performances in BO using a popular representative model from each group, the random forest-based Lolo model (frequentist) and the latent variable Gaussian process model (Bayesian).
We examine the efficacy of the two models in the optimization of mathematical functions, as well as properties of structural and functional materials, where we observe performance differences as related to problem dimensionality and complexity.
By investigating the machine learning models' predictive and uncertainty estimation capabilities, we provide interpretations of the observed performance differences.
Our results provide practical guidance on choosing between frequentist and Bayesian uncertainty-aware machine learning models for mixed-variable BO in materials design.
\end{abstract}
\begin{document}
\flushbottom

\maketitle
\thispagestyle{empty}

\section*{Introduction}
The goal of materials design is to identify materials with desired properties and performance that meet the demands of engineering applications, from among the vast composition--structure design space, which is challenging due to the highly nonlinear underlying physics and the combinatorial nature of the design space.
The traditional trial-and-error approach usually involves many experiments or computations for the evaluation of materials properties, which can be expensive and time-consuming and thus cannot keep pace with the growing demand.
To accelerate materials development with low cost, data-driven adaptive design methods have recently been applied\cite{noh2019solid, akshay-nanocomp, MITs-BO, danial2021acta}.
The adaptive design process starts with small data, selectively adds new samples to guide experimentation/computation, and navigates towards the global optimum.
The key to adaptive materials design is an efficient policy for searching the chemical/structural design space for the \emph{global} minimum, such that new samples (material designs) are selected based on existing knowledge.
Classical metaheuristic optimization methods, such as simulated annealing (SA) and genetic algorithm (GA), select new design samples based on nature-inspired stochastic rules.
However, these methods require many design evaluations, and thus lack cost efficiency, which limits their applicability in materials design.

In contrast, Bayesian Optimization (BO) \cite{BO_review} represents a generalizable and more efficient adaptive design approach.
Starting from a small set of known designs, BO iteratively fits ML models that predict the performance and quantify the uncertainty associated with unseen designs, and then selects new designs to be evaluated in the next iteration based on an acquisition function.
BO methods have demonstrated capabilities in the design optimizations of a diversity of materials, including piezoelectric materials\cite{Yuan2018BaTiO3}, catalysts\cite{Tran2018intermetal}, phase change memories\cite{Kusne2020bal}, and structural materials\cite{Tala2018bma}.
Through these successful cases, BO has shown its versatility, as well as its high efficiency under a limited budget for design evaluation.
Thus it has the potential of being an essential component of data-driven design automation, benefiting materials researchers who are not experts in data science.

Acquisition functions guide the sampling process in BO.
Commonly used acquisition functions, such as expected improvement (EI)\cite{EIacq}, take into account both exploitation (pursuing a better objective) and exploration (reducing uncertainty).
While exploitation is modulated by the ML model's prediction, exploration relies on the estimation of uncertainty in the predicted response for the unsampled sites.
Therefore, uncertainty-aware machine learning (ML) models, i.e., ML models with uncertainty quantification (UQ), play a central role in BO.
Various approaches have been developed to equip ML models with the UQ capability, which we will further discuss in the following section.

However, the mixed-variable problems, i.e., when design variables include both numerical and categorical ones, pose additional challenges to uncertainty-aware ML, and are ubiquitous in materials design.
The design variables in materials design tasks typically include processing, composition, and structure information.
Some design variables such as process type (e.g., hydrothermal or sol-gel), element choice (e.g., Al or Fe), and lattice type (e.g., \emph{fcc} or \emph{bcc}) are categorical, while others such as annealing temperature, stoichiometry, and lattice parameters, are numerical.
For BO methods to be generally applicable to these diverse design representations, uncertainty-aware ML models must be able to handle mixed-variable inputs.

In this work, we first examine the methods for quantifying uncertainty in ML models and contrast their fundamental differences from a theoretical perspective.
Based on this, we focus on two representative uncertainty-aware mixed-variable ML models that involve frequentist and Bayesian approaches to uncertainty quantification, respectively, and conduct a systematic comparative study of their performances in BO, with an emphasis on materials design applications.
Based on the results, we characterize the relative suitability of frequentist and Bayesian approaches to uncertainty-aware ML as related to problem dimensionality and complexity.
Our contribution is two-fold: 
\begin{itemize}
    \item Outline of the suitability of Bayesian and frequentist uncertainty-aware ML models depending on the characteristics of problems;
    \item Identification of key factors that result in the performance difference between Bayesian and frequentist approaches.
\end{itemize}
We anticipate this study will assist researchers in physical sciences who use BO, providing practical guidance in choosing the most appropriate model that suits their purpose.

\section*{Uncertainty-Aware Machine Learning}
\subsection*{Uncertainty in Machine Learning Models}
Uncertainty is ubiquitous in predictive computational models. Even if the underlying physics is deterministic, uncertainty still exists due to the insufficiency of knowledge.
Many efforts have been devoted to quantifying uncertainties of physics-based computational models\cite{arendt2012identify, acs-uq-ml, dft-uq, npj-uq-am} in science and engineering.

Unlike a physics-based model, the prediction of a data-driven model builds upon observations or previous data. Uncertainty in the prediction arises from (1) lack of data, (2) imperfect fit of the model to the data, and (3) intrinsic stochasticity.
These collectively form the metamodeling uncertainty\cite{metamodel-uq}, which reflects the discrepancy between the data-driven model's prediction and the response given by the physics-based model in unsampled regions.
BO's sampling strategy is aimed at reducing the metamodeling uncertainty (exploration) and improving the objective function value (exploitation) by querying certain new samples.
To this end, it is desired to have uncertainty-aware ML models, for which the metamodeling uncertainty can be quantified.

\subsection*{Frequentist and Bayesian Uncertainty Quantification}
Several UQ techniques have been adopted to attain uncertainty-aware ML. Here, we group them into two broad categories: frequentist and Bayesian.
The frequentist approach obtains uncertainty estimation through various forms of resampling: in general, a series of models $\{\hat{f}_i(\bm{x})\}_{i=1}^n$ are fitted with different subsets of training data or hyperparameters, then the prediction variability at an unsampled location is estimated from the variance among these models' predictions:
\begin{equation}
    \hat{s}^2(\bm{x}) = Var\left(\hat{f}_1(\bm{x}), \dots, \hat{f}_n(\bm{x})\right),
\end{equation}
where $Var(\cdot)$ can represent any variance estimate using frequentist statistics, potentially involving noise or bias correction terms.
Commonly used resampling techniques include Monte Carlo, Jackknife, Bootstrap, and their variations \cite{lookman-uq-jap}.
In particular, uncertainty estimation using ensemble models \cite{ensemble-uq} or disagreement/voting of multiple models\cite{hanneke2014theory} are also examples of the resampling approach.

The frequentist UQ approach has been adopted in combination with various ML models, including random forests \cite{shaker2020aleatoric, mentch2016quantifying}, boosted trees \cite{malinin2021uncertainty}, and deep neural networks\cite{uq-dl, Du_Barut_Jin_2021, hirschfeld2020uncertainty}.
As this approach is generally applicable regardless of the type of ML model, it is frequently coupled with ``strong learners'', i.e., the models that are capable of accurately fitting highly complex and non-stationary functions.

Instead of requiring a series of models, the Bayesian UQ approach treats the true model as a random field, and infers its posterior probability from the prior belief and observed data \cite{BO_review} to estimate the uncertainty.
A prominent example is Gaussian process (GP) \cite{williams2006gaussian}.
When modeling the data $(\bm{X},\bm{y})$, a GP model views the observed response $\bm{y}$ as the true response $\bm{f}$ plus random noise.
It assumes that the response $\bm{f}$ at different input locations are jointly Gaussian, i.e., $\bm{f}|\bm{X} \sim \mathcal{N}(\bm{\mu}, \bm{K})$.
The covariance matrix $\bm{K}$ is inferred from the similarity between inputs using a kernel function. 
It also takes into account the noise that may be present in observations by assuming $\bm{y} | \bm{f} \sim \mathcal{N}(\bm{f}, \sigma^2\bm{I})$.
For example, the radial basis function (RBF) kernel
\begin{equation}
    K\left(y(\bm{x}), y(\bm{x}')\right)=\sigma^2\exp\left\{ -\sum_{i}\omega_i(x_i-x_i')^2 \right\}
\end{equation}
uses Euclidean distance metric and assigns Gaussian correlations \emph{a priori}, with global variance $\sigma^2$ and correlation parameters $\omega_i$, to be learned via maximum likelihood estimation (MLE) in model training.
The prediction derived from a GP model includes both the mean and the variance, thus providing a measure of metamodeling uncertainty.

Besides GP, examples of ML methods with Bayesian-style uncertainty estimation include Bayesian linear regression and generalized linear models, Bayesian model averaging \cite{park2010bayesian}, and Bayesian neural networks \cite{bayes-nn}.
When the posterior is not available in analytical form, estimation of the posterior requires probabilistic sampling techniques such as Markov Chain Monte Carlo, whose high computational cost limits its application to various ML models\cite{mcmc-bayes}.
Therefore, the Bayesian UQ approach is often adopted in specially designed ML models that have an analytical form posterior, such as GP.

\subsection*{Uncertainty Quantification in Mixed-variable Machine Learning}
For mixed-variable problems, uncertainty quantification of ML models becomes more complicated.
In early developed ML methods, categorical variables are mostly handled by ordinal or one-hot encoding\cite{eslch2}.
Ordinal encoding assigns integer labels for each category; such encoding assumes ordered relations among categories, thus limiting its applicability.
One-hot encoding represents a categorical variable $t_i$ that takes value from categories (often referred to as levels) $\{\ell_1, \ell_2, \ldots, \ell_J\}$ with a binary vector
\begin{equation}
    \bm{c}_i = [c_i^{(1)},\dots,c_i^{(J)}], c_i^{(j)}=\mathbbm{1}_j,
\end{equation}
where $\mathbbm{1}_j$ is an indicator function, i.e., when $t_i=\ell_j$ only the $j$-th element of $\bm{c}_i$ equals 1, and others equal 0.
This encoding, however, assumes symmetry between all categories (the similarity between any two categories is equal\cite{hase2021gryffin}), which is generally not true.  

In recent years, some methods have been proposed for uncertainty-aware ML in the mixed variable scenario.
Lolo\cite{Lolo}, for example, is an extension of the random forest (RF) model.
As an ensemble of decision trees, RF has native support for mixed-variable problems.
Uncertainty is quantified by calculating variance at any sample point from the predictions of the decision trees with bias correction.

GP models in the original form are uncertainty aware; however, they have problems handling categorical variables.
The covariance matrix is inferred from the similarity between inputs characterized by a distance metric.
But the aforementioned representations cannot represent the distances between categories.
The latent variable Gaussian process (LVGP) model\cite{LVGP, LVGP-BO} solves this problem by mapping each categorical variable $t_i$ into a continuous-variable latent space, where each level $\ell_j$ of $t_i$ is represented by a vector $\bm{z}_i = [z_i^{(1)}(j), \ldots, z_i^{(q)}(j)]$, where $q$, the dimensionality of latent space, is usually 2. The RBF kernel then becomes
\begin{equation}
    K\left(y(\bm{x}, \bm{t}), y(\bm{x}', \bm{t}')\right) = \sigma^2 \exp\left\{ -\sum_i\omega_i(x_i-x_i')^2 - \sum_i \left\|\bm{z}(t_i) - \bm{z}(t_i')\right\|_2^2 \right\},
\end{equation}
where $\|\cdot\|_2$ is the $L^2$ norm.
Like other parameters, locations of latent vectors are obtained via MLE during model training.
With the latent variable representation, the categories are not required to be ordered or symmetric, and their correlations are inherently estimated via distances in the mapped latent space.
The latent variable configuration in the latent space also indicates the effects of different levels of a categorical variable on the response, thus making the model interpretable \cite{LVGP}.
There are extensions of LVGP\cite{wang-biglvgp, da-lvgp} that allow utilizing large training data, physical knowledge, as well as kernels other than RBF that are suitable for fitting functions with different characteristics.
In this work, the vanilla LVGP is used in comparative studies.

\subsection*{Related Comparative Studies}
Some related studies have compared the performances of a variety of UQ techniques in materials design applications.
For example, Tian et al. \cite{lookman-uq-jap} compared four uncertainty estimators among the frequentist ones in materials property optimization.
Liang et al. \cite{liang2021benchmark} conducted a benchmark study of BO for materials design using GP and RF models with different acquisition functions.
However, existing studies focus on BO where input variables are numerical, whereas practical materials design problems are often mixed-variable problems.
The performance of ML methods using frequentist or Bayesian UQ techniques in BO under different circumstances involving categorical variables is not clear.
In particular, the efficacy of the two approaches in the materials design context has not yet been examined.
We hope to fill the gap in this study.

\section*{Results and Discussion}
\begin{figure}[ht]
    \centering
    \includegraphics[width=0.8\textwidth]{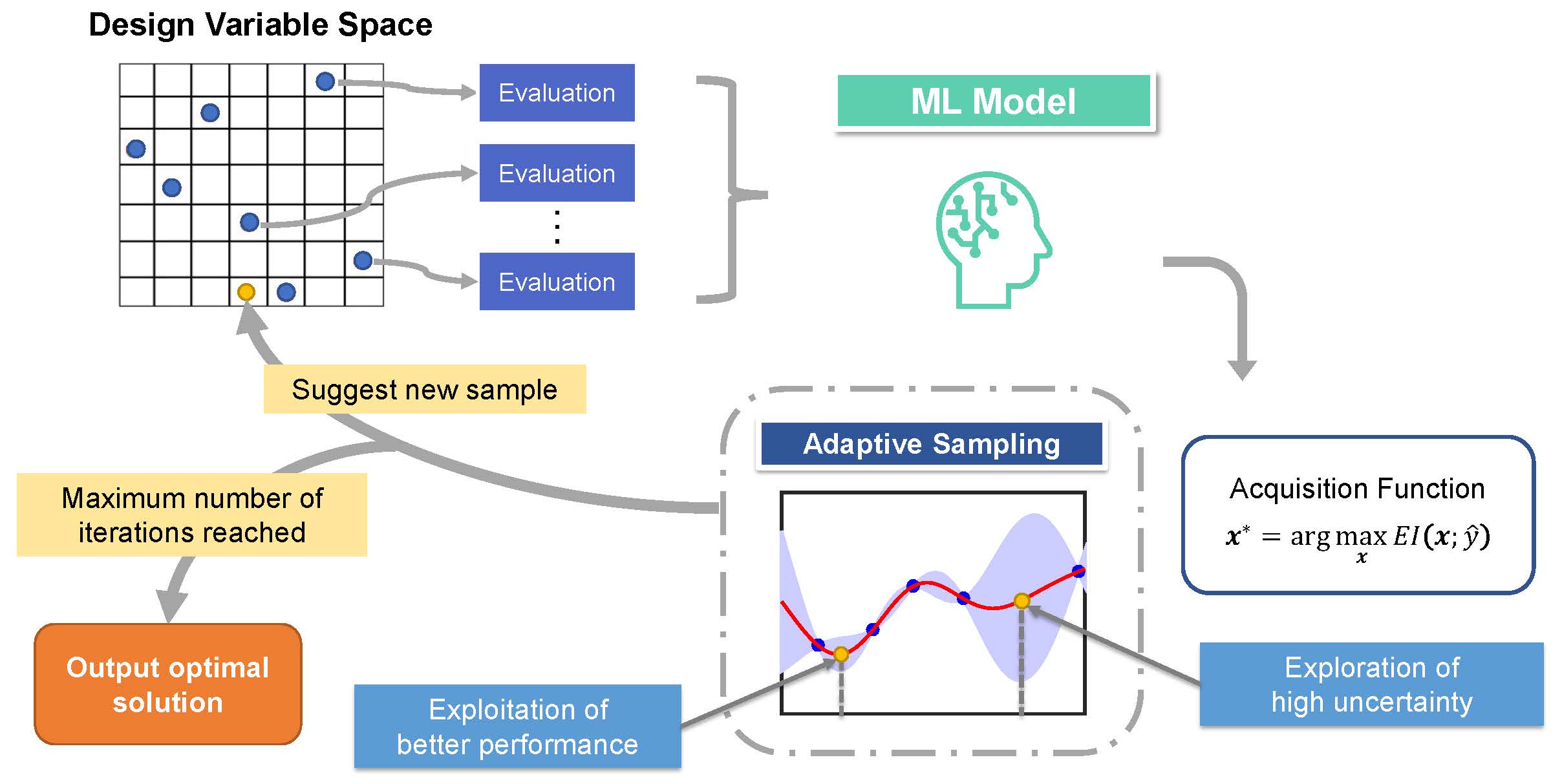}
    \caption{Schematic of Bayesian Optimization framework. An ML model is fitted to the known input--response data, and predicts the response for unevaluated inputs with uncertainty. The acquisition function is calculated from the prediction, guiding the selection of new input(s) to evaluate. The process iterates to find the optimal response.}
    \label{fig:BO_frame}
\end{figure}

To examine and compare the performances of Bayesian Optimization using the two ML models (denoted LVGP-BO and Lolo-BO for conciseness), we tested them on both synthetic mathematical functions and materials property optimization problems.
Our comparative study is conducted using a modular BO framework (illustrated in Figure \ref{fig:BO_frame}; details of implementation are in Methods), in which both LVGP and Lolo can serve as the ML model.

We use this BO framework to search for the minimum value of any function $y(\bm{v})$, where the input variables $\bm{v}=[\bm{x}, \bm{t}]$ consists of numerical variables $\bm{x}$ and/or categorical variables $\bm{t}$, and the response $y$ is a scalar.
The BO performances are compared in two aspects, accuracy and efficiency.
Accuracy relates to the ability to find the optimal objective function value.
We record the complete optimization history for every test case, so that accuracy can be compared by looking at the optimal objective values observed at any time in the optimization process.
Efficiency, on the other hand, is characterized by the rate of reducing the objective function.
In application scenarios such as materials design, the design evaluation (experimentation or physics-based simulation) is often very time-consuming, in comparison, the time for fitting ML models and calculating acquisition functions is negligible.
Thus, when comparing efficiency, we focus on the time in terms of iteration number instead of actual computational time.
The BO method capable of converging to the global optimum in fewer iterations is favored under these metrics.
In the following part, we introduce the experimental settings and present the results for each test problem.

\subsection*{Demonstration: Mathematical Test Functions}
We first present test results of minimizing mixed-variable mathematical functions selected from an online library\cite{simulationlib}.
For functions that are originally defined on a continuous domain, we convert some of their arguments to be categorical for testing purposes.
As these are white-box problems, we can investigate the BO methods' performance under different problem characteristics, as well as the factors influencing the performance.

\subsubsection*{Low-dimensional Simple Functions}  \label{sec:low-d-smooth}
\begin{figure}[ht]
    \centering
    \includegraphics[width=\textwidth]{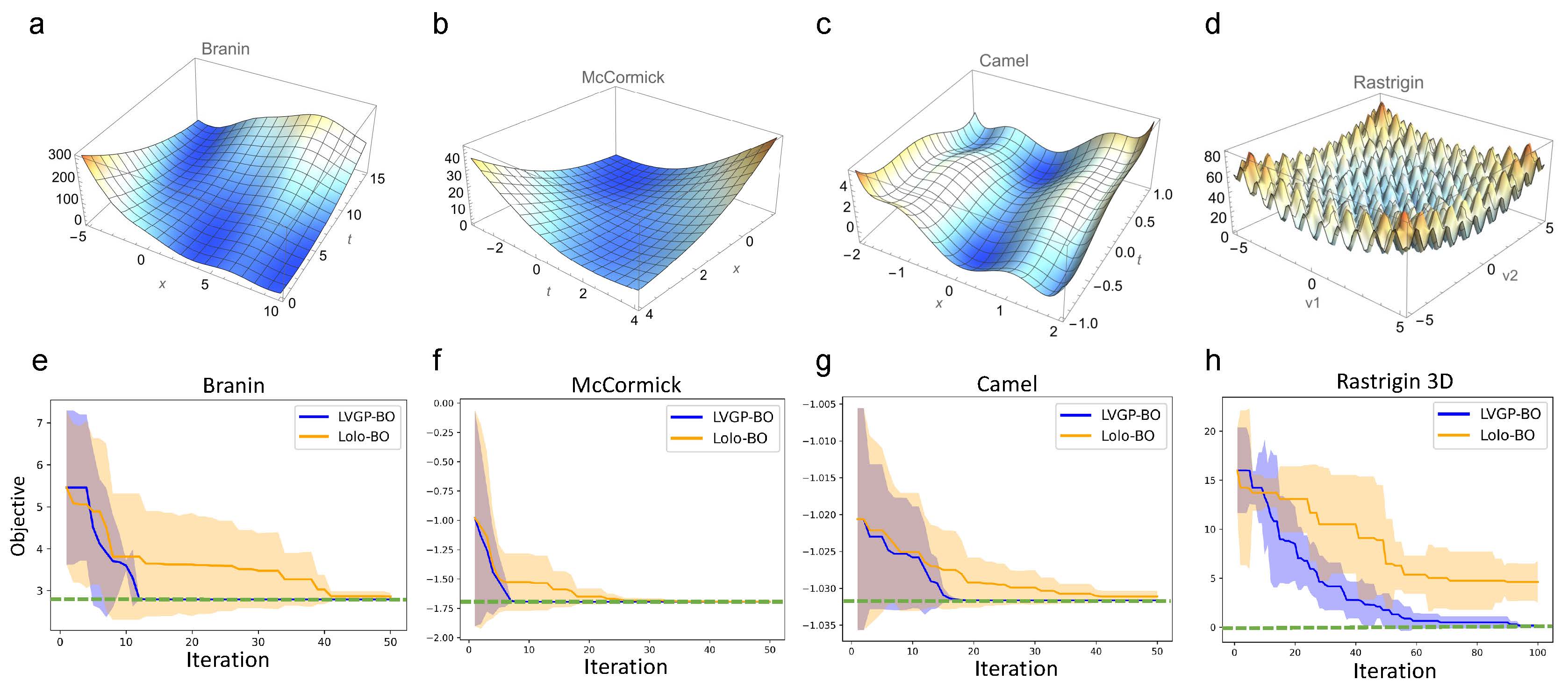}
    \caption{a--d. Visualization of the Branin, McCormick, Camel, and Rastrigin functions in 2-dimensional (2D) continuous form. e--h. Optimization histories across replicates for the Branin, McCormick, Camel, and Rastrigin functions. These plots show the minimal objective function values observed at every iteration. Solid lines represent the median among replicates; shaded areas show plus/minus one MAD. The green dashed lines mark the global minimum value of each function.}
    \label{fig:low-dim}
\end{figure}

In the first test case, we use the Branin function, which has two input variables and relatively smooth behavior. We modify its definition as follows:
\begin{equation}
    f(x,t)=\left(t-\frac{5.1}{4\pi^2} x^{2}+\frac{5}{\pi} x-6\right)^{2}+10\left(1-\frac{1}{8\pi}\right) \cos(x)+10,
\end{equation}
where $x\in[-5,10]$ is a numerical variable, and $t$ is categorical, with categories corresponding to values $\{0, 5, 10, 15\}$.
To provide an intuitive sense of its behavior, we visualize the function in Figure \ref{fig:low-dim}a.
Lolo-BO and LVGP-BO are used respectively to minimize the modified Branin function, starting with 10 initial samples.
We repeat this 30 times with different random initial designs for each replicate, and the optimization histories across replicates are shown in Figure \ref{fig:low-dim}e.
To compare the overall performance and robustness of LVGP-BO and Lolo-BO, we show both the median objective value $\tilde{y}$ and the scaled median absolute deviation $\mathrm{MAD}=\mathrm{median}\left(\left|y-\tilde{y}\right|\right)/0.6745$ at every iteration.

Another low-dimensional, simple test function is the McCormick function (visualized in Figure \ref{fig:low-dim}b), in the following modified form:
\begin{equation}
    f(x,t) = \sin(x+t) + (x-t)^2 - 1.5x + 2.5t +1,
\end{equation}
where $x\in[-1.5,4]$, and $t$'s categories correspond to integer values $\{-3, -2, \dots, 4\}$. The initial sample size and number of replicates are the same as described above; optimization histories are shown in Figure \ref{fig:low-dim}f.
From the optimization history plots, we observe that for both test functions, LVGP-BO converges to the global minimum in fewer iterations, thus showing better efficiency.

\subsubsection*{Low-dimensional Complex Functions}
We then test LVGP-BO and Lolo-BO in optimizing low-dimensional complex functions (definitions are provided in Methods), in this case, rugged functions with several local and/or global minimums.
The Six-Hump Camel function (Figure \ref{fig:low-dim}c):
\begin{equation}
    f(x,t) = \left(4-2.1x^2+\frac{x^4}{3}\right)x^2 + xt + (-4+4t^2)t^2,
\end{equation}
with $x\in[-2,2]$ and $t\in \{\pm 1, \pm 0.7126, 0\}$, is optimized in 30 runs, each starting from initial samples of size 10. As Figure \ref{fig:low-dim}g shows, both LVGP-BO and Lolo-BO converge to the global minimum, while LVGP leads to faster convergence.

We also test optimizing the Rastrigin function, which has more local minimums (as shown in Figure \ref{fig:low-dim}d):
\begin{equation}
    f(\bm{v}) = 10d + \sum_{i=1}^d [v_i^2 - 10\cos(2\pi v_i)],
\end{equation}
where $d$ is the adjustable dimensionality. We set $d=3$, with two numerical variables $x_{1,2}=v_{1,2} \in [-5.12,5.12]$, and one categorical variable $t = v_3 \in \{-5, -4, \dots, 5\}$.
As Figure \ref{fig:low-dim}h shows, in optimizing this highly multimodal function, LVGP-BO shows more performance superiority: it approaches the global minimum at around 60 iterations and eventually converges to the global minimum, while Lolo-BO does not.

\subsubsection*{High-dimensional Functions}  \label{sec:hi-d}
\begin{figure}[ht]
    \centering
    \includegraphics[width=0.8\textwidth]{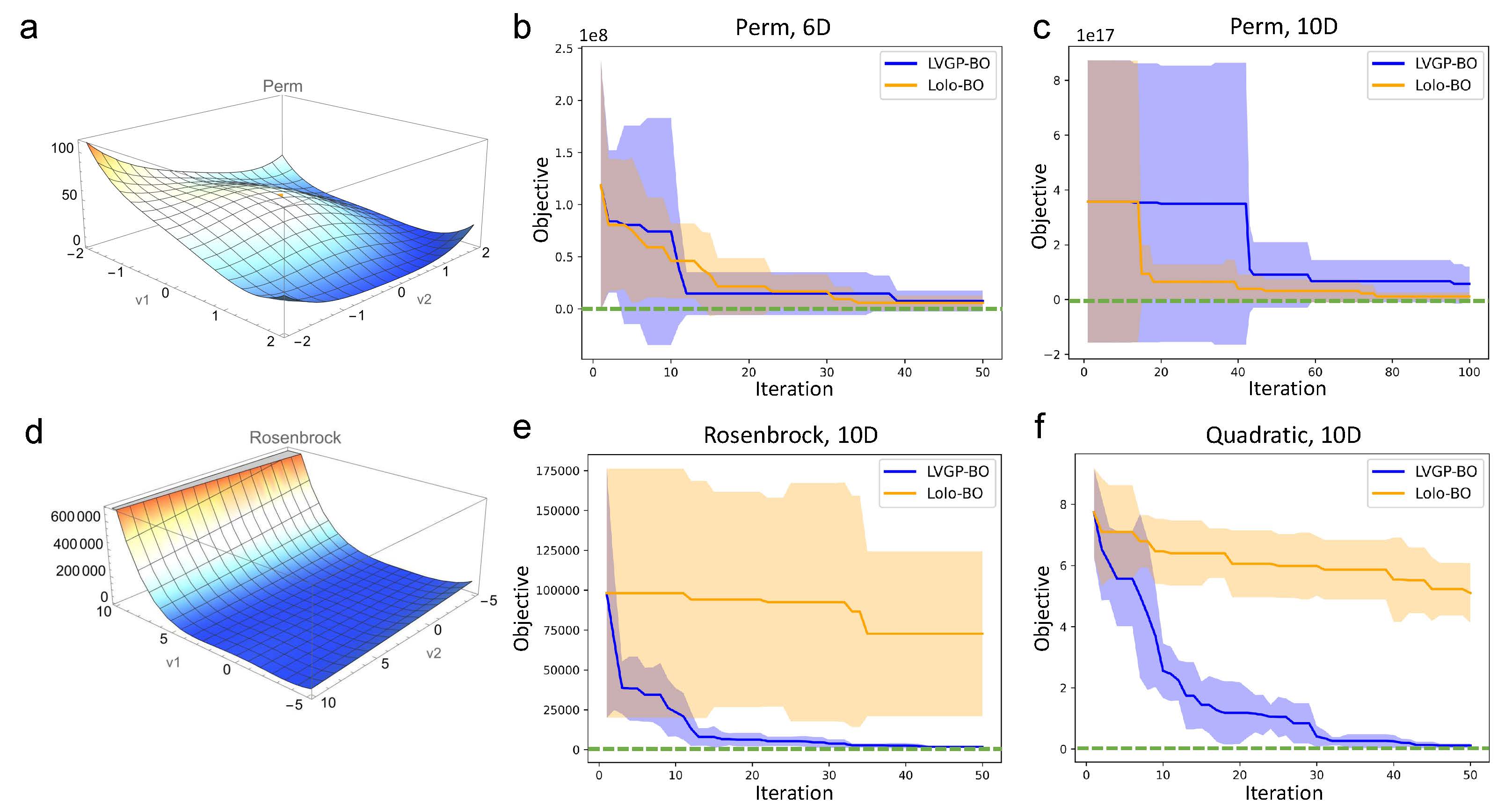}
    \caption{a. Visualization of the Perm function in 2D continuous form. b--c. Optimization histories for the 6D and 10D Perm functions. d. Visualization of the Rosenbrock function. e. Optimization history for the 10D Rosenbrock function. f. Optimization history for the 10D quadratic function.}
    \label{fig:quad_perm}
\end{figure}

Moving beyond low dimensionality, we compare the two BO methods on a series of high-dimensional functions. We are optimizing the Perm function (whose behavior in 2D is shown in Figure \ref{fig:quad_perm}a):
\begin{equation} \label{eq:perm}
    f(\bm{v}) = \sum_{i=1}^{d}\left(\sum_{j=1}^{d}\left(j^{i}+0.5\right)\left(\left(\frac{v_{j}}{j}\right)^{i}-1\right)\right)^{2},
\end{equation}
the Rosenbrock function (whose behavior in 2D is shown in Figure \ref{fig:quad_perm}d):
\begin{equation}
    f(\bm{v})=\sum_{i=1}^{d}\left[100\left(v_{i+1}-v_{i}^{2}\right)^{2}+\left(v_{i}-1\right)^{2}\right],
\end{equation}
and a simple quadratic function $f(\bm{v}) = \sum_{i=1}^d v_i^2$, where $d$ denotes the dimensionality.

For the Perm function, we used both low- and high-dimensional settings: (1) 6-dimensional (6D), with $t = v_6 \in \{-4,1,6\}$ and $x_{1,\dots,5} = v_{1,\dots,5} \in [-6,6]$. In this case, the degrees of freedom $D_f=7$.
(2) 10-dimensional (10D), with $t_1=v_6\in\{-4,1,6\}$, $t_2=v_7\in\{-8,-3,2,7\}$, $t_3=v_8\in\{-6,1,8\}$, $t_4=v_9\in\{\pm 3, \pm 9\}$, $t_5=v_{10}\in\{0, \pm 5, \pm 10\}$, and $x_{1,\dots,5} = v_{1,\dots,5} \in [-10,10]$. $D_f=19$ for this function.
10 replicates are run for each test, starting from initial samples of size 20 for 6D and 50 for 10D. Observations are that, in the 6D test case, LVGP-BO and Lolo-BO show close efficiencies (Figure \ref{fig:quad_perm}b).
Whereas in the 10D case, both BO methods have difficulties optimizing the function and get stuck for more than 20 iterations (Figure \ref{fig:quad_perm}c); Lolo-BO displays better convergence rate and final minimum objective value.

The Perm function is complex because of its non-convexity, and more significantly, its erratic behavior at the domain boundary: the function value is growing nearly exponentially near the boundary.
We test BO of the 10D Rosenbrock and quadratic functions to investigate the influence of high dimensionality, without the erratic complexity.
The Rosenbrock function is also non-convex, but is well-behaved at the domain boundary.
We define its input variables as following: numerical variables $x_{1,\ldots,5}=v_{1,\ldots,5} \in [-5, 10]$, categorical variables $t_1=v_6\in\{-4,1,6\}$, $t_2=v_7\in\{-3, 1, 5, 9\}$, $t_3=v_8\in\{-5, 1, 7\}$, $t_4=v_9\in\{-2, 1, 4, 7\}$, $t_5=v_{10}\in\{\pm 3, \pm 1, 5\}$, which make $D_f = 19$.
The quadratic function is convex and well-behaved. It takes five categorical variables $t_{1,\dots,5} = v_{6,\dots,10} \in \{0, \pm 1, \pm 2\}$ and five numerical variables $x_{1,\dots,5} = v_{1,\dots,5} \in [-2,2]$, with $D_f = 25$.
As Figure \ref{fig:quad_perm}e--f shows, both BO methods make progress in descending the function value towards the optimum, while LVGP-BO has a considerably faster convergence rate.
Through these, we find that when the dimensionality of the problem is high, convexity influence the comparison between LVGP-BO and Lolo-BO similarly to the low-dimensional situation.
For the three 10D functions, Lolo-BO displays consistent behavior of making slow progress; whereas when the function is ill-behaved near the domain boundary, the efficiency of LVGP-BO decreases.
In the Supplementary Information (SI), we present additional test cases, which support the findings as well.

\subsubsection*{What Determines BO Performance?}
We seek explanations for LVGP-BO and Lolo-BO's performance differences from two aspects: fitting accuracy and uncertainty estimation quality.
For BO to successfully locate the global optimum, the ML model does not need to fit the response function accurately everywhere, but the accuracy near the optimum matters.
This accuracy can be improved with new sample acquisition guided by uncertainty.
In regions that are far from optimal, the quantity of data and the resulting prediction accuracy only need to be sufficient to confidently rule the region out as a promising design region, which is why uncertainty quantification is important. 

\begin{figure}[ht]
    \centering
    \includegraphics[width=0.8\textwidth]{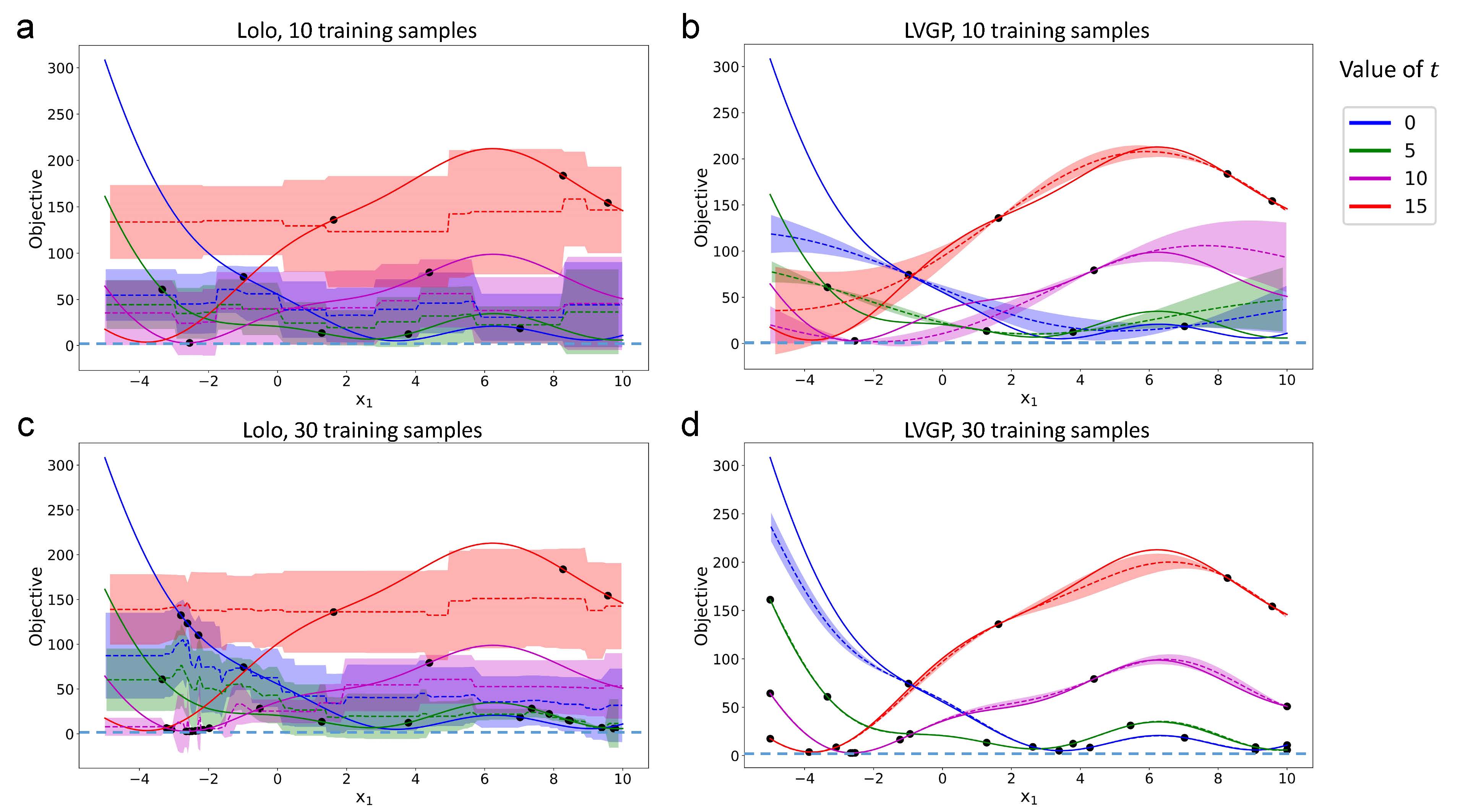}
    \caption{Illustration of the behaviors of LVGP and Lolo fitting the Branin function, with 10 and 30 samples. Each panel plots $f(x,t)$ versus $x$ for four fixed values of $t$, while different colors of curves indicate levels of $t$. Solid lines represent the true function value, black dots are sample points in the training set, dashed lines are the predicted mean value, and shaded areas show the uncertainty estimation (plus/minus one standard deviation). The global minimum function value is marked by dashed lines.}
    \label{fig:branin_fit}
\end{figure}

We select the Branin function as a representative, generate 10 initial samples following the same as described above, run 20 iterations of BO to acquire 20 more samples, and fit LVGP/Lolo models to the samples of sizes 10 and 30.
In Figure \ref{fig:branin_fit} we show the behaviors of LVGP and Lolo in fitting the mixed-variable Branin function.
With a small training set, LVGP can attain better prediction of the function compared to Lolo; moreover, its uncertainty quantification assigns low uncertainty in the vicinity of known observations and high uncertainty in the regions where data are sparse.
These enable well-directed sampling in the less explored regions, hence promoting the model to ``learn'' the target function efficiently in regions where it matters most, i.e., in the vicinity of the optimum.
In SI, we show the sampling sequences of Lolo-BO and LVGP-BO optimizing the Branin function to illustrate the difference between LVGP’s and Lolo’s uncertainty quantification and their effects on sample selection.

\begin{figure}[ht]
    \centering
    \includegraphics[width=0.8\textwidth]{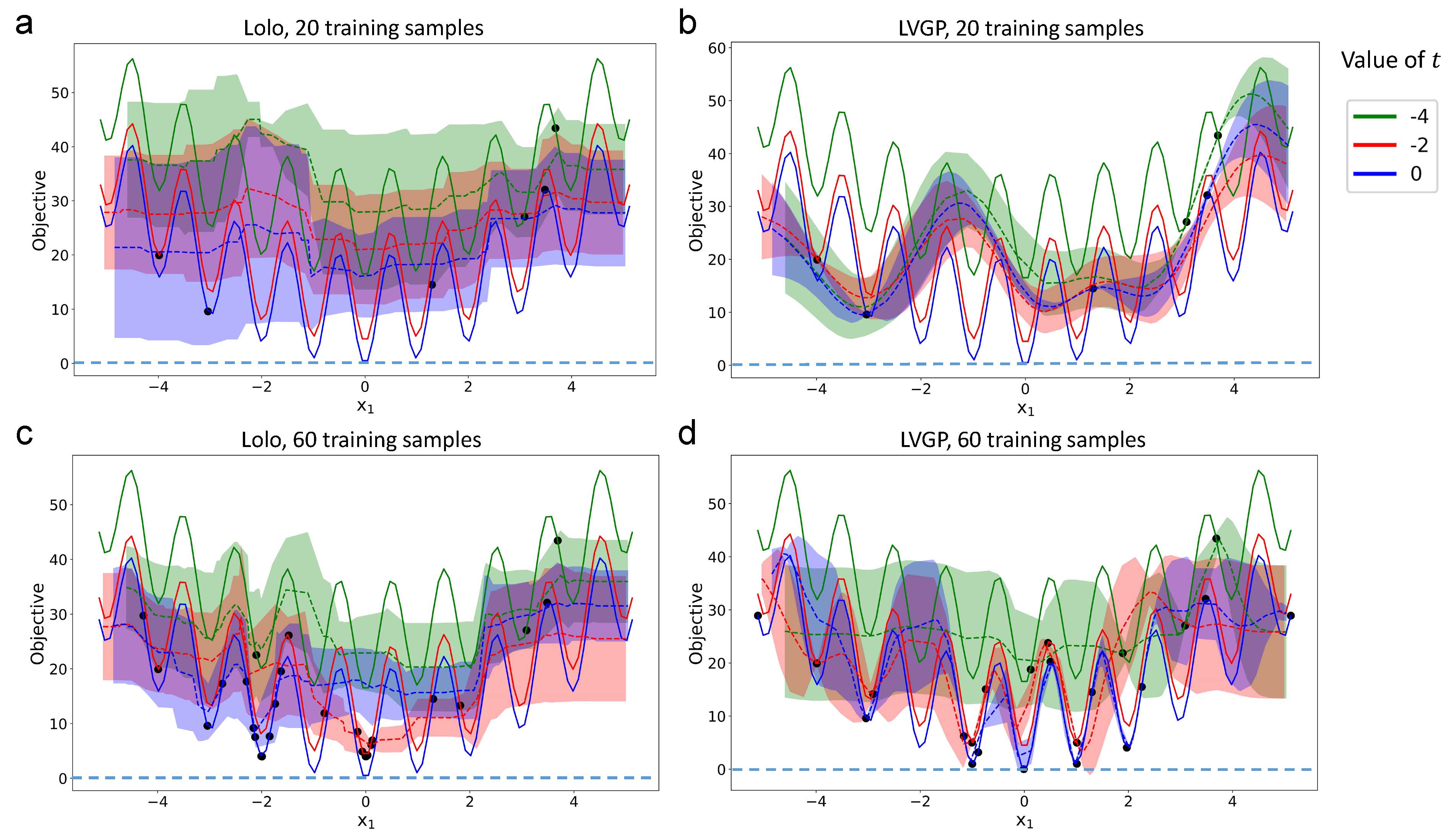}
    \caption{Illustration of the behaviors of LVGP and Lolo fitting the 2D Rastrigin function, with 20 (initial) and 60 (BO expanded) samples. For clarity, we only show three levels of categorical variable $t$ out of eleven in total.}
    \label{fig:rastrigin_fit}
\end{figure}

We conducted a similar fitting test for the Rastrigin function in 2D and show the results in Figure \ref{fig:rastrigin_fit}.
For this more complex function, both methods fail to attain a good fit with 20 initial samples.
Despite this, LVGP gives a better estimation of uncertainty in that it assigns higher uncertainty at the regions with sparser data points, which effectively guides the model towards a better fit.
With the training sample size increased to 60, however, LVGP can fit the fluctuating function more accurately than Lolo (compare panels c and d), especially in the regions close to optimum, i.e., where the function values are low.

\begin{figure}[ht]
    \centering
    \includegraphics[width=\textwidth]{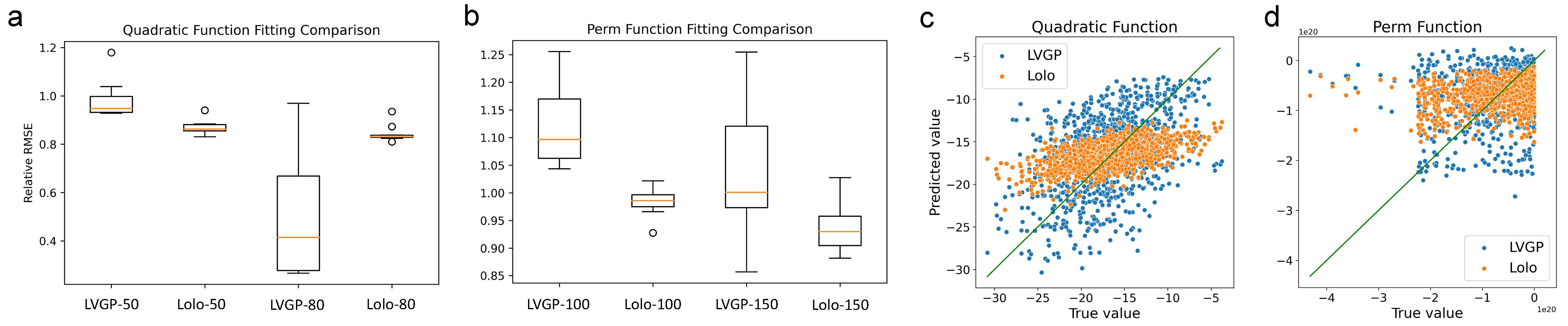}
    \caption{a--b. Relative RMSE of fitting quadratic and Perm functions, using LVGP and Lolo models with varying training sample sizes. Boxplots show results from 10 different randomly selected training sample sets. c--d. Regression plots showing the true value (horizontal axis) and ML model-predicted value (vertical axis) for quadratic function (training size 50) and Perm function (training size 100), with green diagonal lines representing accurate predictions.}
    \label{fig:hi-d-fit}
\end{figure}

We next extend this fitting and UQ comparison to high-dimensional cases.
Training samples are generated from the 10-dimensional quadratic function and the Perm function (Equation \ref{eq:perm}), respectively, then LVGP and Lolo models are tested to accurately fit the samples.
At high dimensions, the previous visualization is no longer feasible. Instead, we adopt the relative root-mean-square error (RRMSE)
\begin{equation}
    RRMSE = \frac{RMSE}{\sigma_y} = \sqrt{\frac{\sum_i (\hat{y}_i-y_i)^2}{\sum_i (y_i-\bar{y})^2}}
\end{equation}
as a metric of fitting quality.
Note that RRMSE is related to another widely used metric, the coefficient of determination $R^2$, through $R^2 = 1 - RRMSE^2$. $RRMSE>1$ can happen when the fitted model is worse than using the response mean as a constant predictor.
For each function, we evaluate the model fitting quality by calculating RRMSE on 1,000 test samples generated independently from the training samples.
Figure \ref{fig:hi-d-fit}a--b shows the RRMSEs across different training sample sets to indicate how well the two models fit the mathematical functions. 
We also show in Figure \ref{fig:hi-d-fit}c--d the deviations of the models' predictions from the true responses, i.e., the prediction errors.

As the figures show, in fitting the relatively simpler quadratic function, Lolo attains a higher quality compared to LVGP at small sample size (50); as the sample size increases to 80, the fitting quality of LVGP improves significantly, whereas the fitting quality of Lolo does not change much.
However, even at a small sample size, LVGP's prediction error for samples with low function values (near optimum) is lower than Lolo's.
With well-directed uncertainty quantification, LVGP-BO can add samples that improve the fitting, thus leading to efficient convergence.

In fitting the 10D Perm function, both functions fail to attain a good RRMSE; the Lolo model fits slightly better than LVGP, and this comparison is not changed as the training sample size increases.
In this case, the dimensionality is too large for the known samples to cover, hence, it is difficult for both ML models to capture the complexity of the Perm function.
Neither model shows dominant fitting accuracy near optimum over another model.
Lolo's slightly better global fitting accuracy enables it to display higher efficiency in BO of the Perm function.

\subsection*{Materials Design Applications}
To assess the performances of two ML models in facilitating materials design, we apply Lolo-BO and LVGP-BO to optimize materials' properties using several existing experimental/computational materials datasets.
We adopt a simple yet generally applicable design representation, using chemical composition as design variables to optimize the properties.
We start with a small fraction of samples randomly selected from the dataset; the evaluation of a sample is imitated by querying its corresponding property from the dataset.
Model fitting and acquisition function follow the same procedure described in previous sections.

\subsubsection*{Moduli of $\rm M_2AX$ Compounds}
The $\rm M_2AX$ materials family \cite{m2ax-review} has a hexagonal crystal structure, in which M, A, and X represent different sites, M and X atoms form a 2-dimensional network with the X atoms at the center of octahedra, while the A atoms connect the layers formed by M and X.
$\rm M_2AX$ compounds display high stiffness and lubricity, as well as high resistance to oxidation and creeping resistance at high temperatures.
These properties make them promising candidates as structural materials in extreme-condition applications such as aerospace engineering. \cite{m2ax-apl, m2ax-jpcm}
For both the capability as a structural material and the manufacturability, elastic properties are of particular importance.
However, elastic properties have nontrivial dependence on composition.
The computational determination of these properties includes the calculation of stresses or energies under several strains using density functional theory (DFT)\cite{m2ax-jpcm}, which is resource-intensive.
Here we demonstrate how the design optimization of $\rm M_2AX$ compounds' elastic properties directly in the composition space may benefit from mixed-variable BO while assessing the performances of Lolo-BO and LVGP-BO.

From Balachandran et al. \cite{M2AX}, we retrieve a dataset that reports Young's, bulk, and shear moduli ($E$, $B$, and $G$, respectively) of 223 $\rm M_2AX$ compounds within the chemical space $\rm M\in\{Sc, Ti, V, Cr, Zr, Nb, Mo, Hf, Ta, W\}$, $\rm A\in\{Al, Si, P, S, Ga, Ge, As, Cd, In,\allowbreak Sn, Tl, Pb\}$, $X\rm \in\{C, N\}$.
The input variable $\bm{v}$ is thus 3-dimensional, with all inputs being categorical.
Since $E$ and $G$ are highly correlated (shown in SI), we choose $E$ and $B$ as target responses and optimize them separately.
Both optimizations start with 30 initial samples and run for 50 iterations, adding one sample per iteration.

\begin{figure}[ht]
    \centering
    \includegraphics[width=\textwidth]{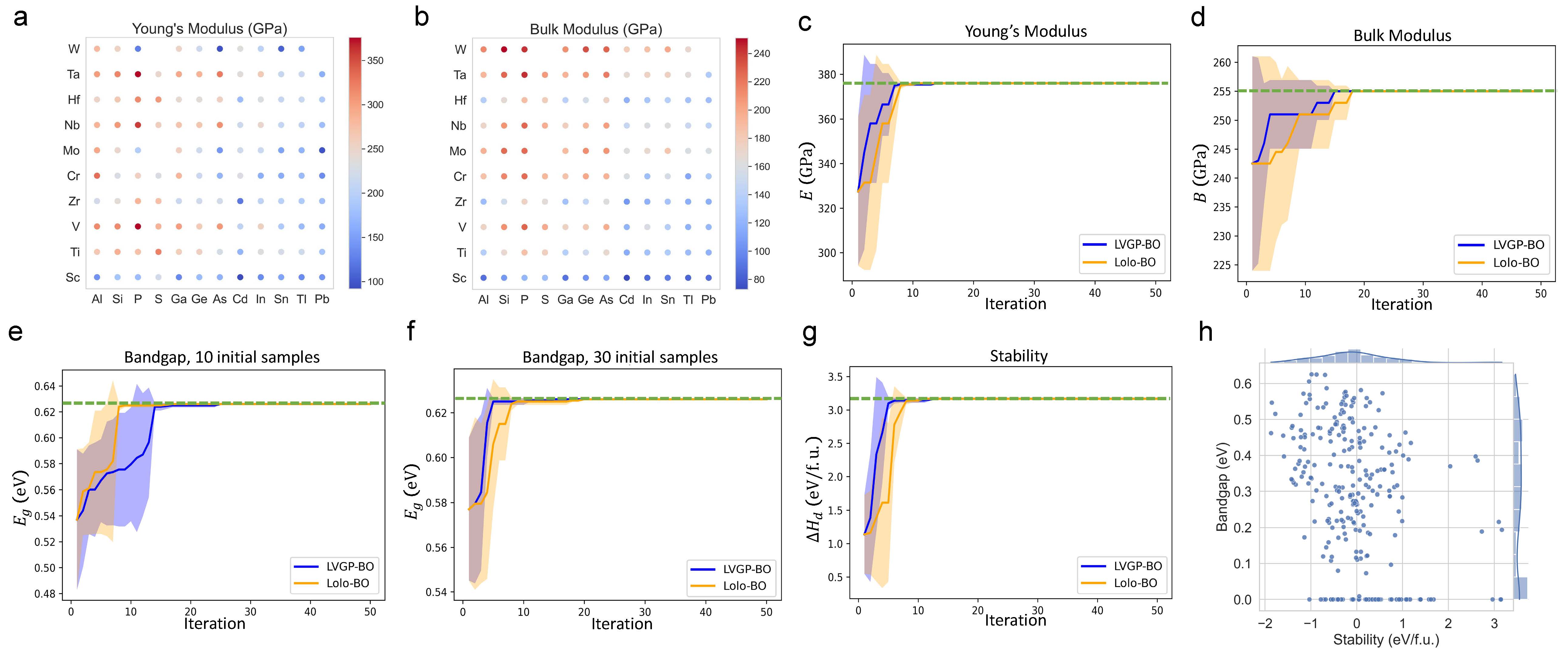}
    \caption{a--b. Distributions of $E$ and $B$ values in the M--A space, fixing $\rm X=C$. c--d. Optimization histories for c. Young's modulus and d. bulk modulus. e--g. Optimization histories of $E_g$ with initial sample sizes 10 and 30, and $\Delta H_d$ with initial sample size of 10. h. Scatter plot of bandgap--stability for all samples in the dataset.}
    \label{fig:matl_opt}
\end{figure}

In Figure \ref{fig:matl_opt}, we show the value distributions and optimization histories for $E$ and $B$.
Both Lolo-BO and LVGP-BO are capable of discovering the material that has optimal modulus within 20 iterations, significantly reducing the required resources as compared to computationally evaluating the whole design space.
Of the two BO methods, LVGP-BO exhibits marginally higher rates of convergence in both tasks.
As observed from a--b, the input--response relations of both $E$ and $B$ are relatively well-behaved, without showing abrupt changes or clusters of values, which favors LVGP-BO's efficiency.
Hence, the results here are consistent with the findings in the mathematical test cases.

\subsubsection*{Bandgap and Stability of Lacunar Spinels}
In another materials design application case, we consider materials having the formula $\rm AM^aM^b_3X_8$ and the lacunar spinel crystal structure\cite{lacunar-spinel}.
Element candidates for the sites are $\rm A\in\{Al, Ga, In\}$, $\rm M^a\in\{V, Nb, Ta, Cr, Mo, W\}$, $\rm M^b\in\{V,Nb,Ta,Mo,W\}$, $\rm X\in\{S,Se,Te\}$.
This is a family of materials that potentially exhibit metal--insulator transitions (MITs)\cite{mits-review}, i.e., electrical resistivity changing significantly upon external stimuli, such as temperature change across a critical temperature.
The MIT property can be leveraged for encoding and decoding information with lower energy consumption compared to current devices\cite{mits-transistor}.
Hence, the $\rm AM^aM^b_3X_8$ materials family shows promise for next-generation microelectronic devices, including neuron-mimicking devices which can accelerate ML\cite{mit-memresitor}.
The origin of the transition is structural distortion triggered by external stimuli, which leads to a redistribution of electrons in the band structure\cite{mits-phys}.
Though the physical mechanism behind MITs is complex, two relevant properties may serve as proxies for the performances of candidate materials.
One is the bandgap of the insulating ground state $E_g$, as a larger $E_g$ generally corresponds to a higher resistivity in the insulating state, and therefore a higher resistivity change ratio upon the phase transition to a metallic phase under the applied field.
Another is the decomposition enthalpy of the material $\Delta H_d$ which is associated with a material's stability.
Stable compounds are more likely to be synthesizable and operable in novel devices.
Therefore, we use these two properties corresponding to their functionality and stability as the target in MIT materials design.

In a dataset collected by Wang et al.\cite{MITs-BO}, a total of 270 combinations of candidate elements are enumerated, for every compound $E_g$ and $\Delta H_d$ calculated from DFT are listed.
Similar to the previous test case, we use the 4-dimensional (categorical) composition as the inputs and optimize two responses $E_g$ and $\Delta H_d$ separately.
Figure \ref{fig:matl_opt}e--g show the results: starting from 10 initial samples, LVGP-BO and Lolo-BO both discover the compound with optimal $\Delta H_d$ efficiently, but are relatively slow in optimizing $E_g$; LVGP-BO shows better efficiency on $\Delta H_d$ while Lolo-BO shows better efficiency on $E_g$.
When we increase the initial sample size to 30, LVGP-BO and Lolo-BO exhibit similar efficiency on $E_g$.

We show the different characteristics of the two responses of the dataset by a scatter plot in Figure \ref{fig:matl_opt}h.
Among the 270 $E_g$ values in the dataset, 56 are zero and others are positive values. These values form a clustered distribution at 0 and make the target function $E_g = f(\bm{v})$ ill-behaved.
Combined with the high-dimensionality, this function becomes challenging for LVGP-BO and Lolo-BO to optimize, as we demonstrated in the high-dimensional numerical examples.
In contrast, $\Delta H_d$ values form a relatively well-behaved target function, hence, LVGP-BO performs better on this task, also in agreement with previous findings.

\subsubsection*{Investigating Machine Learning Performance}
\begin{figure}[ht]
    \centering
    \includegraphics[width=\textwidth]{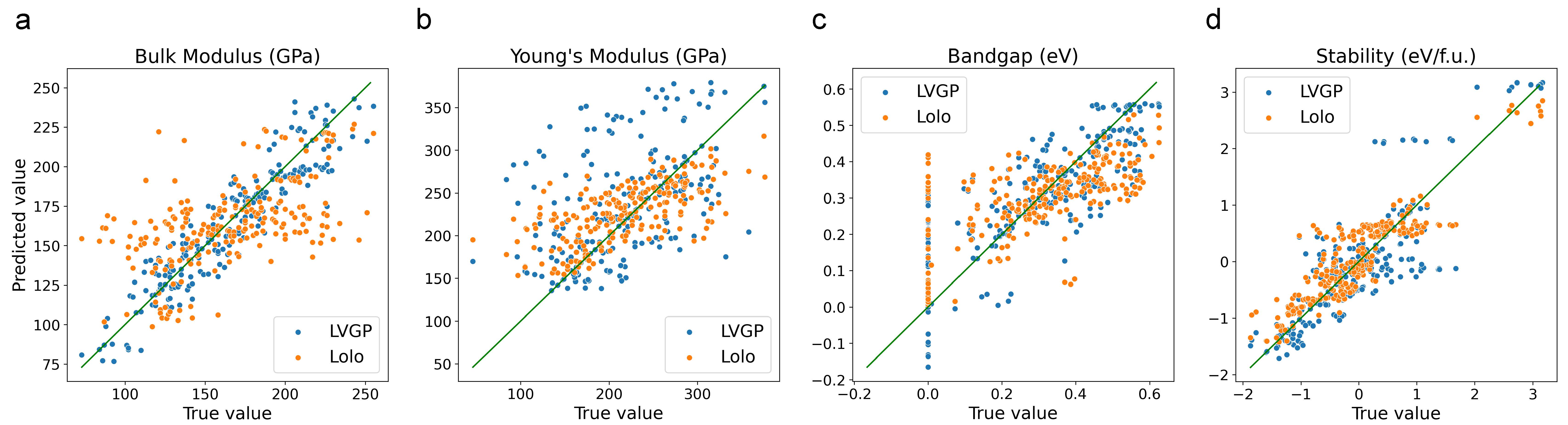}
    \caption{Regression plots for ML models trained on 30 samples for materials properties: a--b. bulk and Young's moduli of $\rm M_2AX$ compounds; c--d. bandgap and stability of lacunar spinels.}
    \label{fig:matl_fit}
\end{figure}

Though the response functions linking materials compositions to properties are black-box functions, for which the exact behaviors are unknown, we investigate the fitting accuracy of two ML models to get a hint of their performance in BO.
The regression plots in Figure \ref{fig:matl_fit} show the test response predictions versus the true response values for the ML models fitted to training data of size 30.
Interestingly, for all four target properties, LVGP shows better prediction errors for the high-performance (larger true response) materials, which are the candidates close to optimum.
Aligning with the findings from mathematical examples, this explains LVGP-BO's edge in efficiency over Lolo-BO.
Another observation worth noting is that, for the lacunar spinels with zero bandgaps, LVGP's predictions contain negative values, while Lolo's do not (Figure \ref{fig:matl_fit}c).
This is because the LVGP model using the RBF kernel tends to yield smooth response function predictions, which accounts for the influence of the clustered behavior of the response function on LVGP-BO's performance.

\section*{Conclusion}
In this study, we examine the fundamental differences between frequentist and Bayesian uncertainty quantification in ML models.
Thereafter, we systematically compare the efficiency and accuracy of BO powered by two representative mixed-variable ML models in mathematical optimization as well as materials design tasks, and investigate the factors influencing BO performances.
In summary, an ML model's fitting accuracy near the optimum and its uncertainty quantification quality are found important to BO.
For low-dimensional problems, the ML models can fit the input--response function relatively easily; even if the function is highly complex, fitting quality can be improved by adding a small number of well-selected samples.
In this case, the quality of UQ becomes the key factor of BO efficiency, where LVGP using Bayesian uncertainty quantification has advantages.
Whereas for high-dimensional problems, if the function is complex, it becomes challenging for ML models to fit the function.
The number of samples required for covering the input space and improving fitting quality also escalates due to the curse of dimensionality.
In this case, better fitting leads to better BO performance, and ML models that are more capable of fitting complex functions (such as random forests) have advantages.

The results and analyses draw a suitability boundary for LVGP-BO and Lolo-BO, and more generally, provide insights for understanding the difference between the two families of uncertainty-aware ML models they represent:
\begin{itemize}
    \item When the design optimization problem is low-dimensional, or high-dimensional but the response is anticipated to be relatively well-behaved, the LVGP model is recommended for BO.
    \item While for high-dimensional problems with a highly ill-behaved response function, we recommend using an ML model that allows higher model complexity (e.g., random forest, neural network) with resampling UQ.
\end{itemize}
The results constitute a supplement to the previous studies covering BO with all numerical variables and guide the model selection in materials design as well as other mixed-variable BO problems.

\section*{Methods}
\subsection*{Bayesian Optimization}
The optimization process starts with initial samples, i.e., an initial set of input variables, and evaluates the responses.
A machine learning model is then fitted with the known input--response data, which assigns for any input $\bm{v}$ a mean prediction $\hat{y}(\bm{v})$ and associated uncertainty (predicted variance) $\hat{s}^2(\bm{v})$.
The model is used to make uncertainty-aware predictions for the unevaluated samples $\mathcal{V}$ (sample pool).
A new sample is selected therefrom based on the expected improvement (EI) acquisition function\cite{EIacq}:
\begin{align}
    \bm{v}^* &= \arg\max_{\bm{v}\in\mathcal{V}} \mathrm{EI}(\bm{v}),\\
    \mathrm{EI}(\bm{v}) &= \mathbb{E}[\max\{0,\Delta(\bm{v})\}] = \hat{s}(\bm{v})\phi\left(\frac{\Delta(\bm{v})}{\hat{s}(\bm{v})}\right)+\Delta(\bm{v})\Phi\left(\frac{\Delta(\bm{v})}{\hat{s}(\bm{v})}\right),
\end{align}
where $\Delta(\bm{v})=y_{\rm{min}}-y(\bm{v})$, the difference between the minimal response value observed so far and the mean prediction of the fitted ML model; $\phi(\cdot)$ and $\Phi(\cdot)$ are the standard normal probability density function (\emph{pdf}) and cumulative distribution function (\emph{cdf}), respectively.
The new sample and corresponding response value are added to the known dataset. This process is repeated iteratively, until the maximum number of iterations or some convergence criterion is reached.

\subsection*{Comparative Experiments}
To compare the performances of Lolo and LVGP, we substitute them as the ``ML Model'' into the framework (Figure \ref{fig:BO_frame}) and run BO for a variety of functions, each time keeping the initial designs the same for BO with Lolo and LVGP.
The initial designs are generated quasi-randomly following a systematic approach: numerical variables are drawn together from a Sobol sequence \cite{sobolseq}; each categorical variable is obtained from shuffling a list where all categories appear equally frequently and at least once.
Since the stochasticity of initial samples influences the optimization process, we run multiple replicates of BO with different random seeds for each test problem.

\subsection*{Uncertainty-Aware ML models}
In these comparisons, we use the open-source implementation of Lolo\cite{lolo-code} in \texttt{Scala} language with the \texttt{Python} wrapper \texttt{lolopy}, and a \texttt{MATLAB} implementation of LVGP, which implements the same algorithm as the open-source package coded in \texttt{R} \cite{lvgp-code}.
For hyperparameters of both models, we use the default settings: For Lolo, the maximum number of trees is set to the number of data points, the maximum depth of trees is $2^{30}$, and the minimum number of instances in the leaf is 1.
For LVGP, we use 2D latent variable mapping and the RBF kernel.
Detailed settings are listed in the open-source packages.

\subsection*{Metrics for Problems Difficulty}
We specify the following metrics to characterize the test problems and BO methods' performance.
The dimensionality of inputs is an important criterion of problem difficulty.
However, in categorical or mixed-variable cases, dimensionality is more than the number of variables. A more useful index considered in this work is the degrees of freedom, which we define as
\begin{equation}
    D_f \equiv \mathrm{dim}(\bm{x}) + \sum_i\left[\#\mathrm{levels} (t_i) -1\right],
\end{equation}
where $\#\mathrm{levels} (t_i)$ yields the number of levels of $t_i$. This quantity takes into account the number of levels for each categorical variable. In other words, high dimensionality may mean ``many levels'' in problems with categorical variables. In the following sections, we follow this definition to categorize problems with $D_f>15$ as high-dimensional, and others as low-dimensional.

Another criterion of difficulty is the complexity of the objective function.
In this work, we view the functions that display the following characteristics as ill-behaved:
\begin{itemize}
    \item rugged: the response fluctuates a lot, resulting in many local minima;
    \item erratic: the response value changes abruptly in certain regions;
    \item clustered: the response takes certain values frequently.
\end{itemize}
These characteristics make a function challenging for ML and global optimization, hence, we refer to them as ``complex'' functions. Conversely, other well-behaved functions, including highly nonlinear ones, are referred to as ``simple'' functions.

\bibliography{refs}

\section*{Acknowledgements}
This work was supported in part by the Advanced Research Projects Agency-Energy (ARPA-E), U.S. Department of Energy, under Grant Number DE-AR0001209. The views and opinions of authors expressed herein do not necessarily state or reflect those of the United States Government or any agency thereof.
The authors thank Bryan L. Horn for assistance in experiments, Alexandru B. Georgescu for providing materials science insights, and Suraj Yerramilli for helpful discussions.

\section*{Author contributions}
H.Z., A.I., and W.C. conceived the experiments. H.Z. conducted the experiments and drafted the manuscript. W.W.C. and D.W.A. contributed to the theoretical analysis. W.C. formulated and supervised the project. All authors reviewed and revised the manuscript.

\section*{Data availability}
The datasets generated during and/or analysed during the current study are available from the corresponding author on reasonable request.

\section*{Additional information}
\paragraph{Competing interests} The authors declare no competing interests.

\end{document}


\maketitle
\section{Uncertainty Quantification in Mixed-Variable ML}
\begin{figure}[H]
    \centering
    \includegraphics[width=0.7\textwidth]{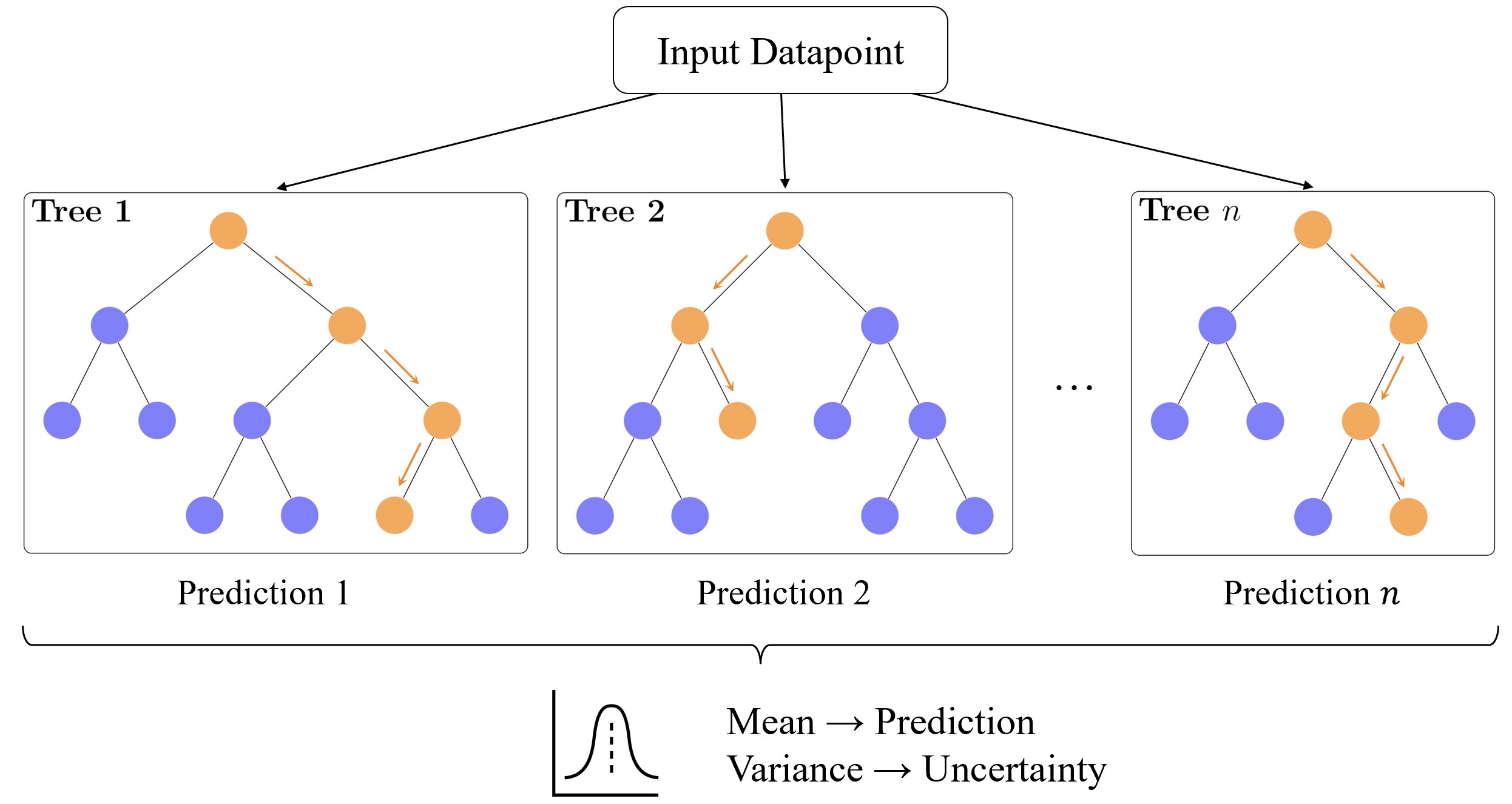}
    \caption{Illustration of the Lolo model. In the training stage, $n$ decision trees are constructed. In the prediction stage, for an input datapoint, each tree gives a prediction. Predicted mean and uncertainty are given by the mean value and variance estimation among $n$ trees' predictions, respectively.}
    \label{fig:rf-graph}
\end{figure}

\begin{figure}[H]
    \centering
    \includegraphics[width=\textwidth]{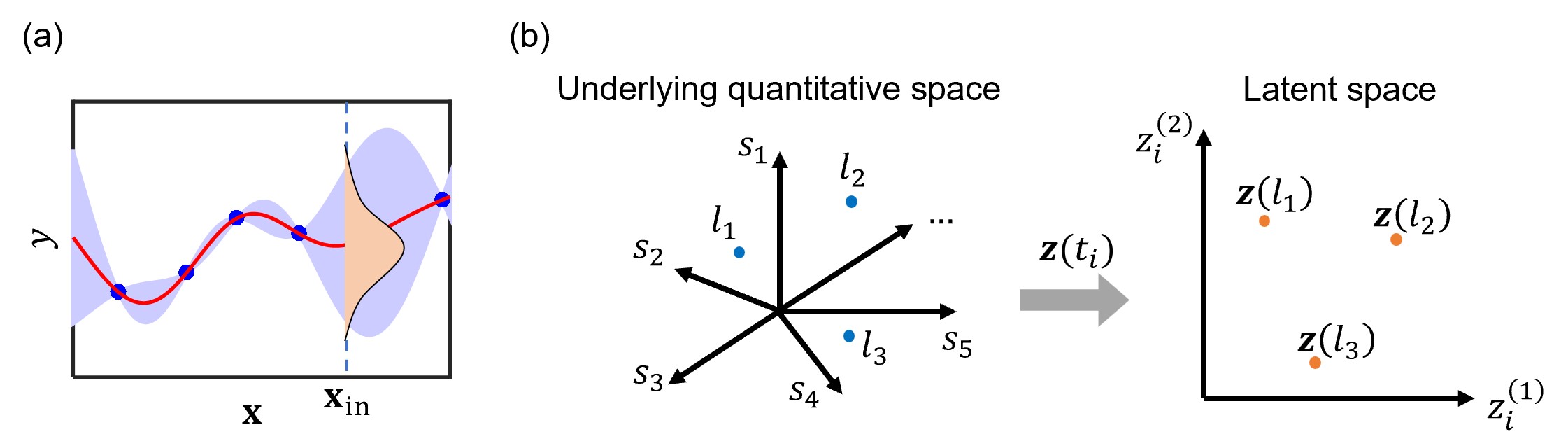}
    \caption{Illustrations of Gaussian Process and LVGP models. (a) GP models assign prior distributions for $y$ at various $\bm{x}$ locations. Given training data, the posterior distribution of $y$ provides uncertainty quantification. (b) LVGP model assumes that levels of a categorical variable are characterized by underlying quantitative factors (represented as $s$). Represented as points in a low-dimensional latent space, the distances between levels reflect their similarities in terms of effect on the response.}
    \label{fig:lvgp-pic}
\end{figure}

Two uncertainty-aware mixed-variable ML models, Lolo and LVGP, are illustrated in Figures \ref{fig:rf-graph} and \ref{fig:lvgp-pic}, respectively.

\section{Mathematical Test Cases}
\begin{figure}[H]
  \centering
  \begin{subfigure}[b]{0.48\textwidth}
    \centering
    \includegraphics[width=\textwidth]{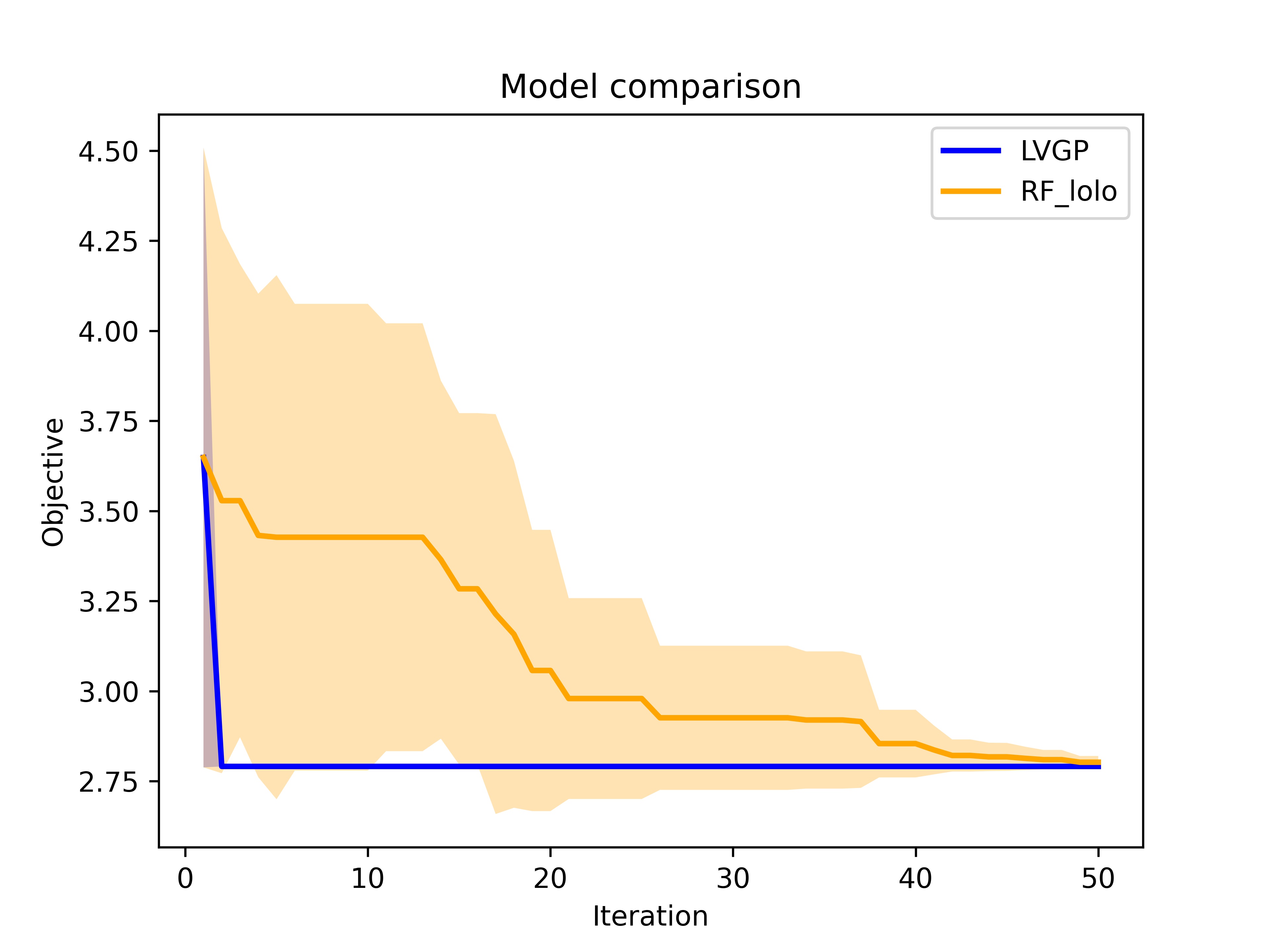}
  \end{subfigure}
  \begin{subfigure}[b]{0.48\textwidth}
    \centering
    \includegraphics[width=\textwidth]{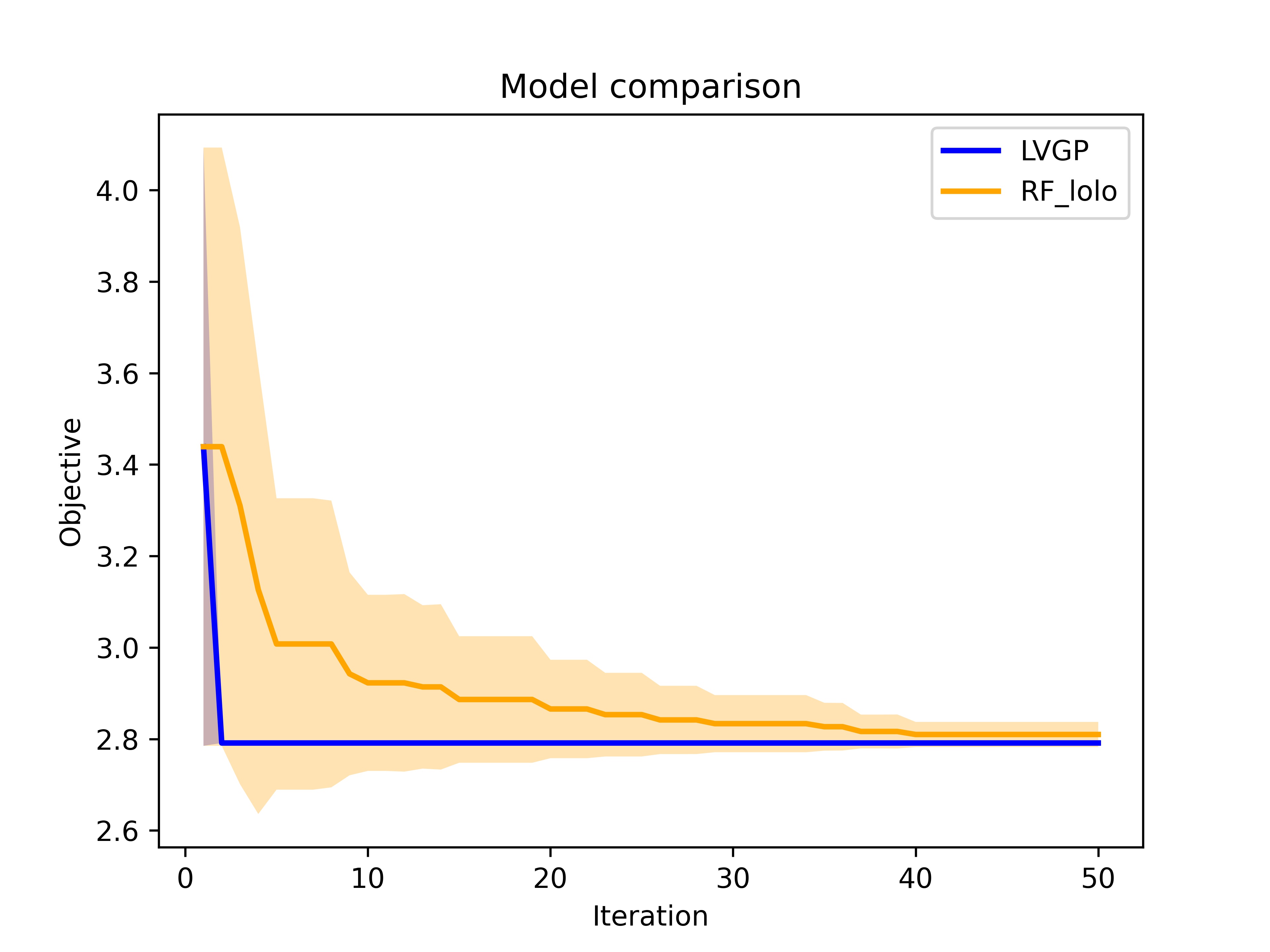}
  \end{subfigure}
  \caption{Optimization history plots of the Branin function with initial sample sizes 50 (left) and 100 (right).}
  \label{fig:branin}
\end{figure}

\begin{figure}[H]
  \centering
  \begin{subfigure}[b]{0.48\textwidth}
    \centering
    \includegraphics[width=\textwidth]{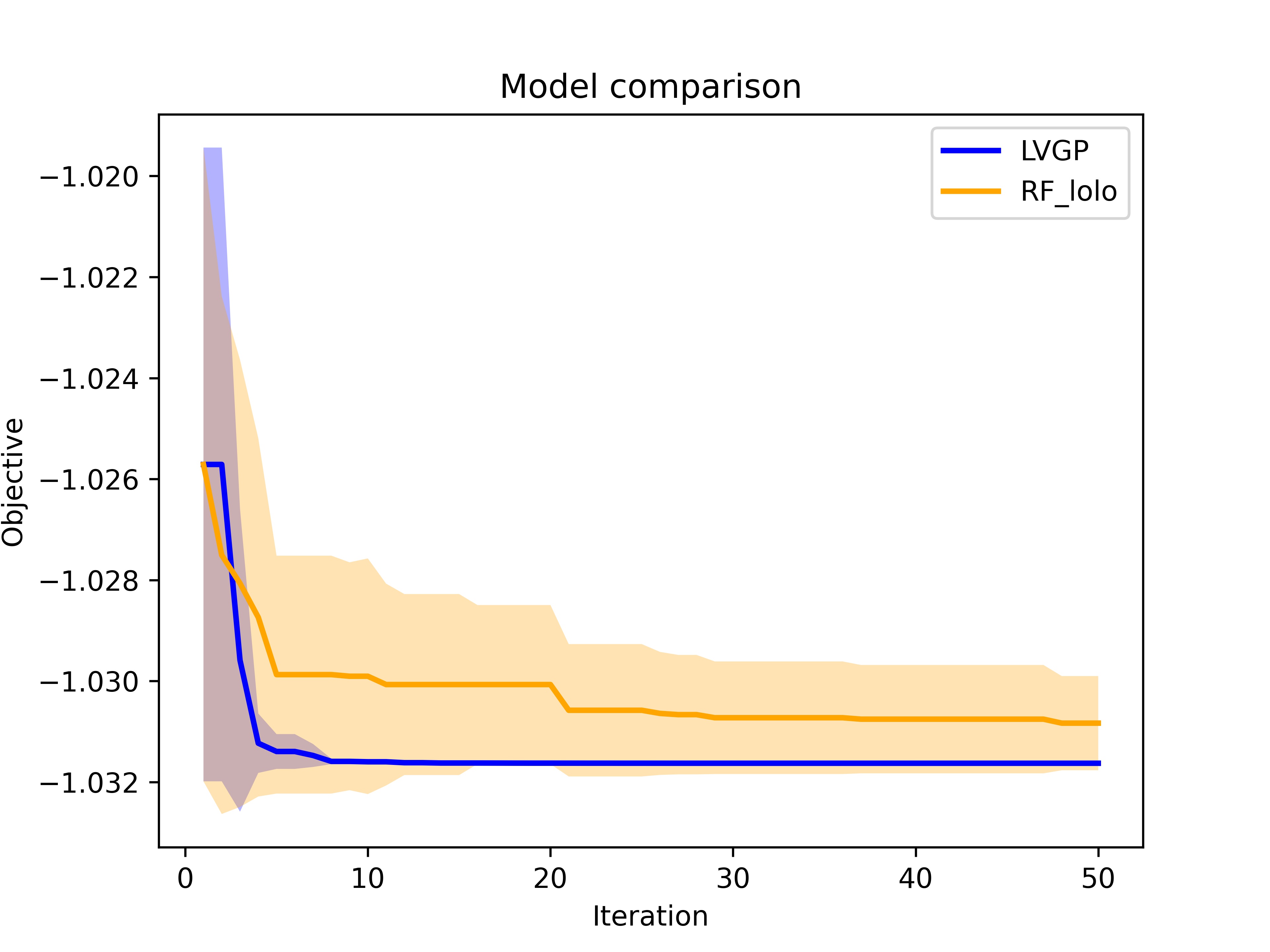}
  \end{subfigure}
  \begin{subfigure}[b]{0.48\textwidth}
    \centering
    \includegraphics[width=\textwidth]{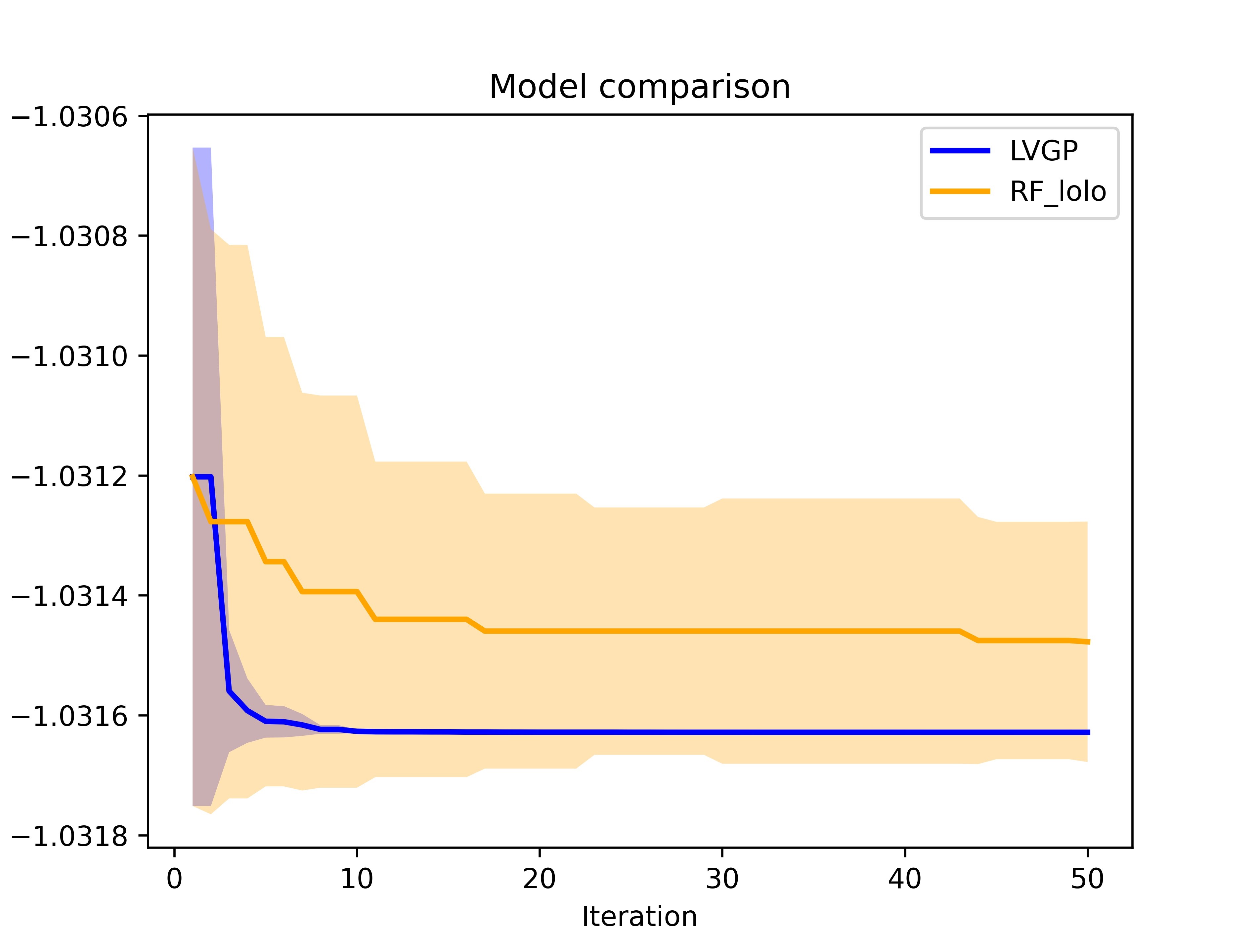}
  \end{subfigure}
  \caption{Optimization history plots of the Camel function with initial sample sizes 50 (left) and 100 (right).}
  \label{fig:camel}
\end{figure}

Bayesian Optimization (BO) results of mathematical test functions included in the main text, with different initial sample sizes: Branin function in Figure \ref{fig:branin}, and Camel function in Figure \ref{fig:camel}.

\begin{figure}[H]
  \centering
  \begin{subfigure}[b]{0.48\textwidth}
    \centering
    \includegraphics[width=\textwidth]{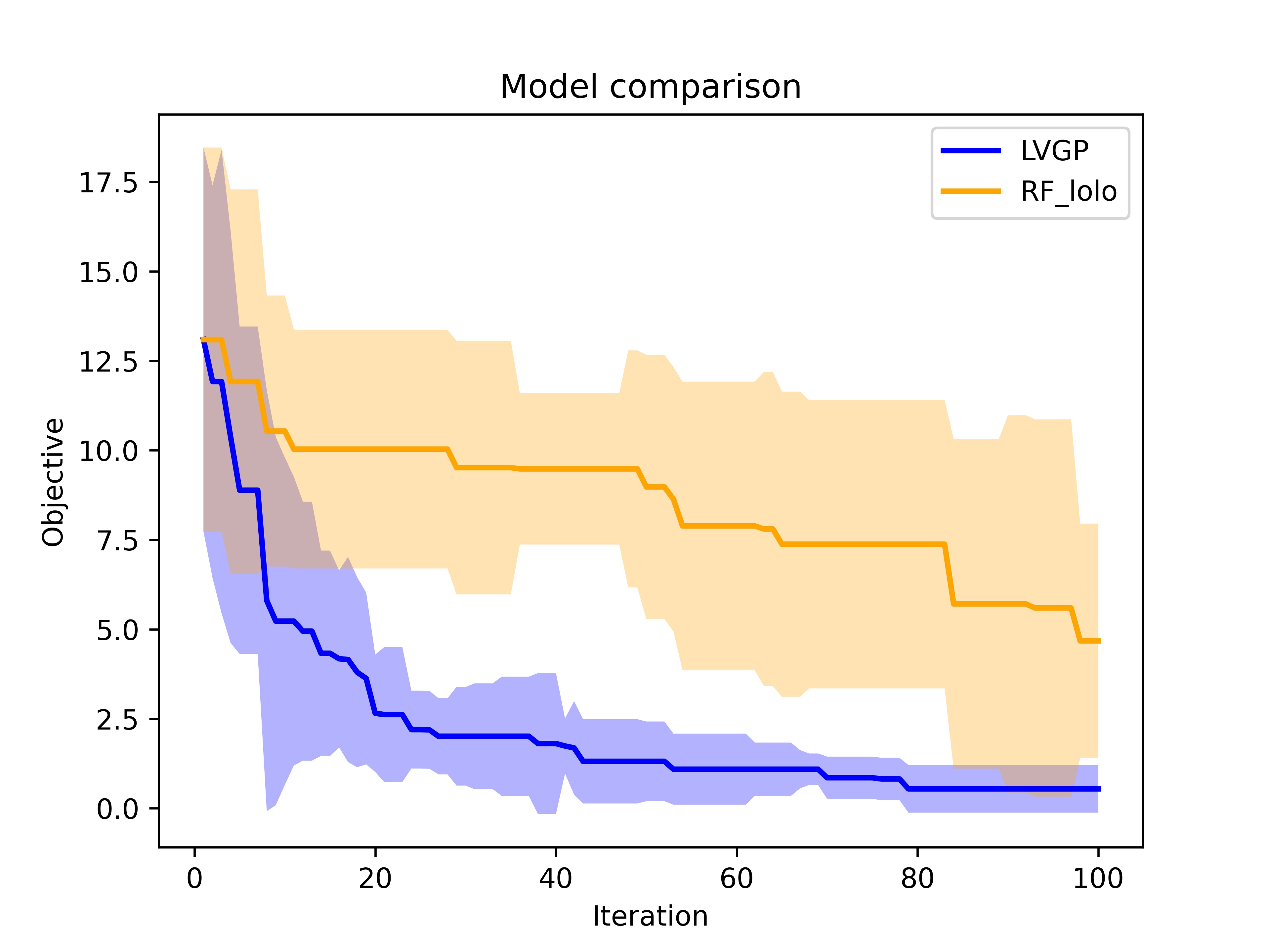}
  \end{subfigure}
  \begin{subfigure}[b]{0.48\textwidth}
    \centering
    \includegraphics[width=\textwidth]{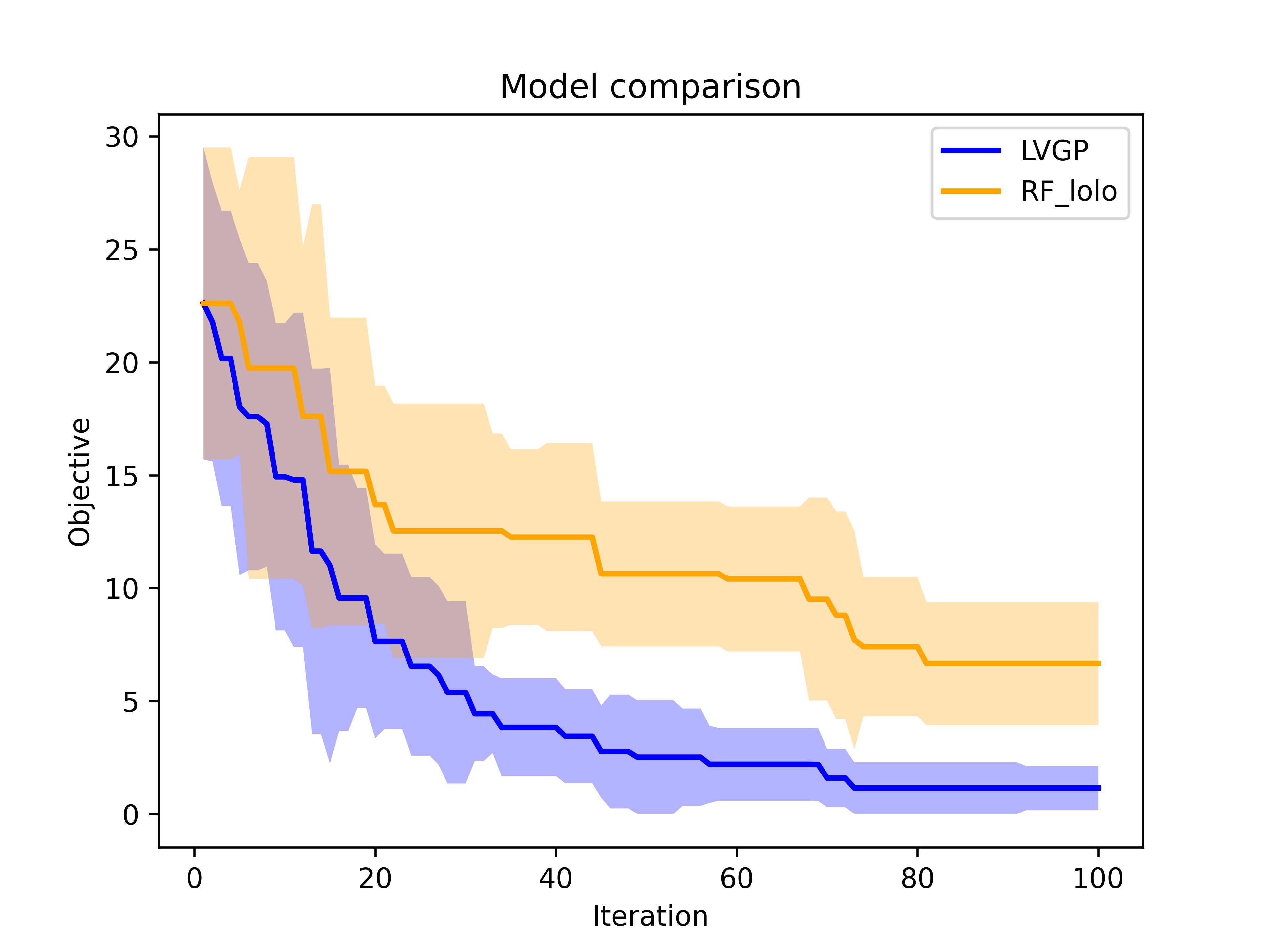}
  \end{subfigure}
  \caption{Optimization history plots of the 3D (left) and 4D (right) Rastrigin functions.}
  \label{fig:rastrigin}
\end{figure}

The Rastrigin function in 3-dimensional (1 categorical) and 4-dimensional (2 categorical) forms, with 30 initial samples. Plots (Figure \ref{fig:rastrigin}) are created from results of 10 runs with different random initial samples.

\begin{figure}[H]
    \centering
    \includegraphics[width=0.5\textwidth]{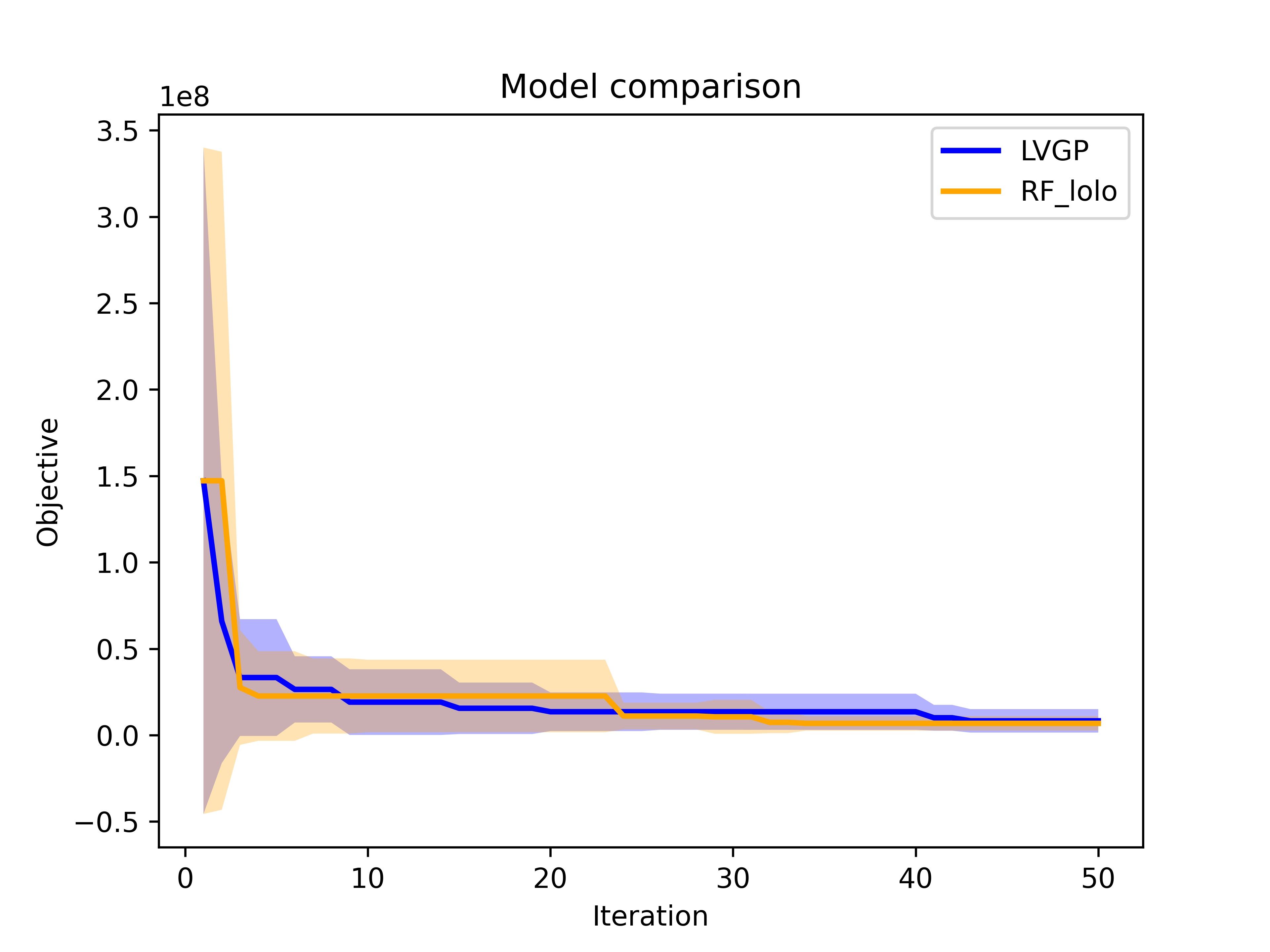}
    \caption{Optimization history plot of the Perm function.}
    \label{fig:perm6}
\end{figure}

The Perm function in 6-dimensional form (1 categorical variable with 3 levels), starting from 80 initial samples. Figure \ref{fig:perm6} is created from 10 runs.

\begin{figure}[H]
  \centering
  \begin{subfigure}[b]{0.48\textwidth}
    \centering
    \includegraphics[width=\textwidth]{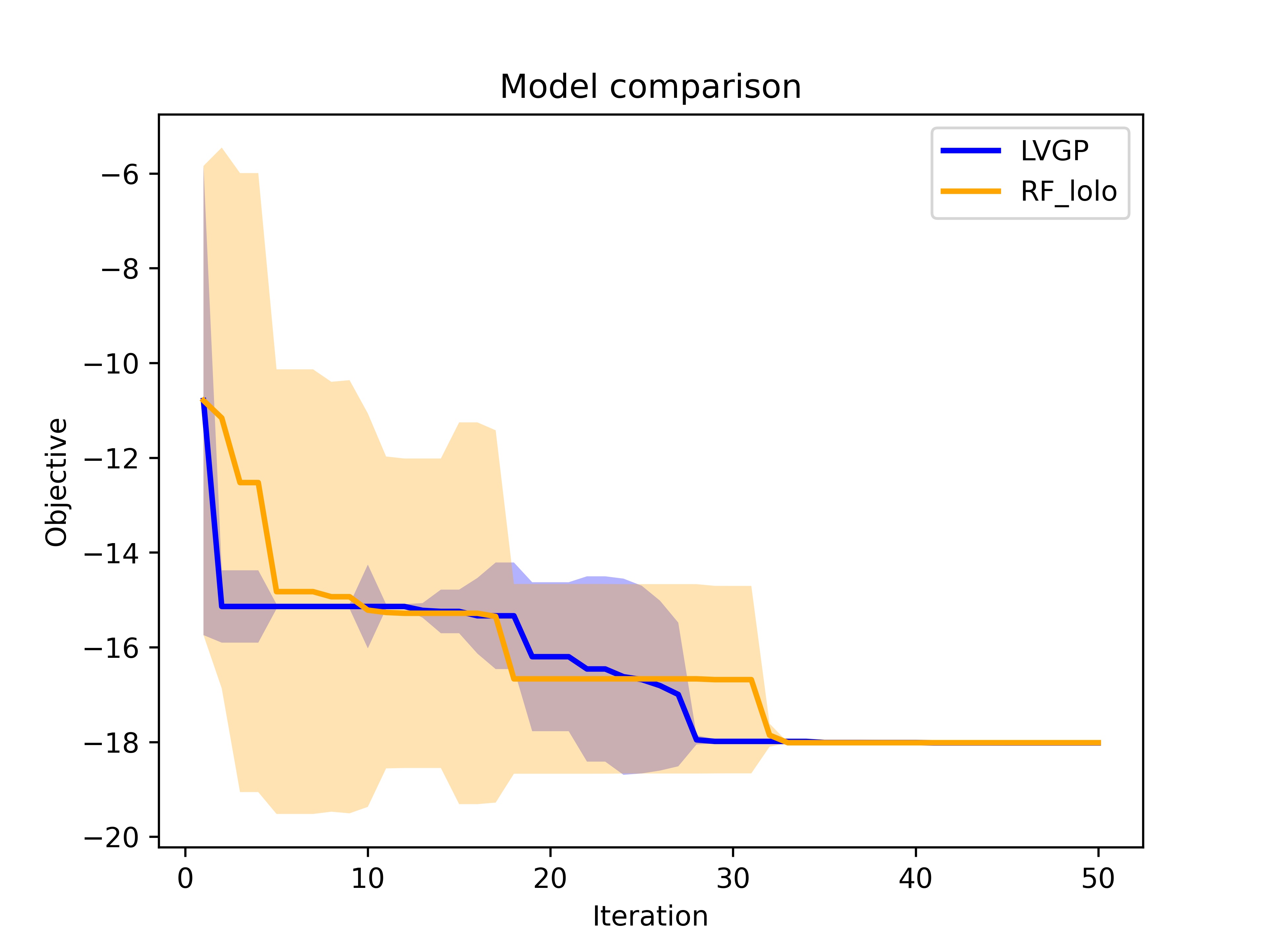}
  \end{subfigure}
  \begin{subfigure}[b]{0.48\textwidth}
    \centering
    \includegraphics[width=\textwidth]{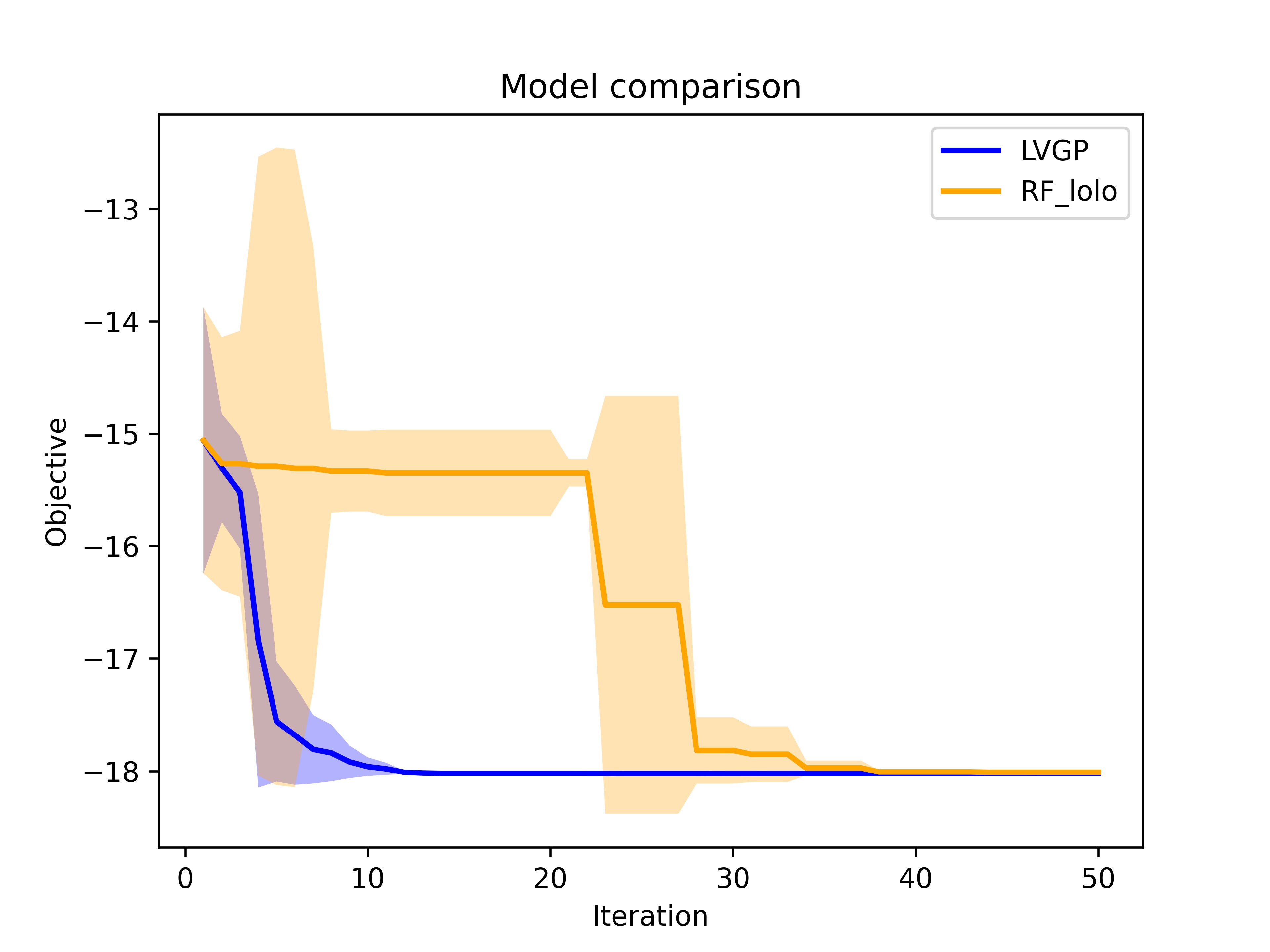}
  \end{subfigure}
  \caption{Optimization history plots of the Holder Table function with initial sample sizes 21 (left) and 50 (right).}
  \label{fig:holder}
\end{figure}

\paragraph{The Holder Table function}
\begin{equation}
    f(x,t)=-\left|\sin \left(x\right) \cos \left(t\right) \exp \left(\left|1-\frac{\sqrt{x^{2}+t^{2}}}{\pi}\right|\right)\right|,
\end{equation}
where $x\in [-10, 10]$, and $t\in\{\pm 10, \pm 9, \dots, 0\}$. Plots (Figure \ref{fig:holder}) are created from 10 runs.

\begin{figure}[H]
    \centering
    \includegraphics[width=0.5\textwidth]{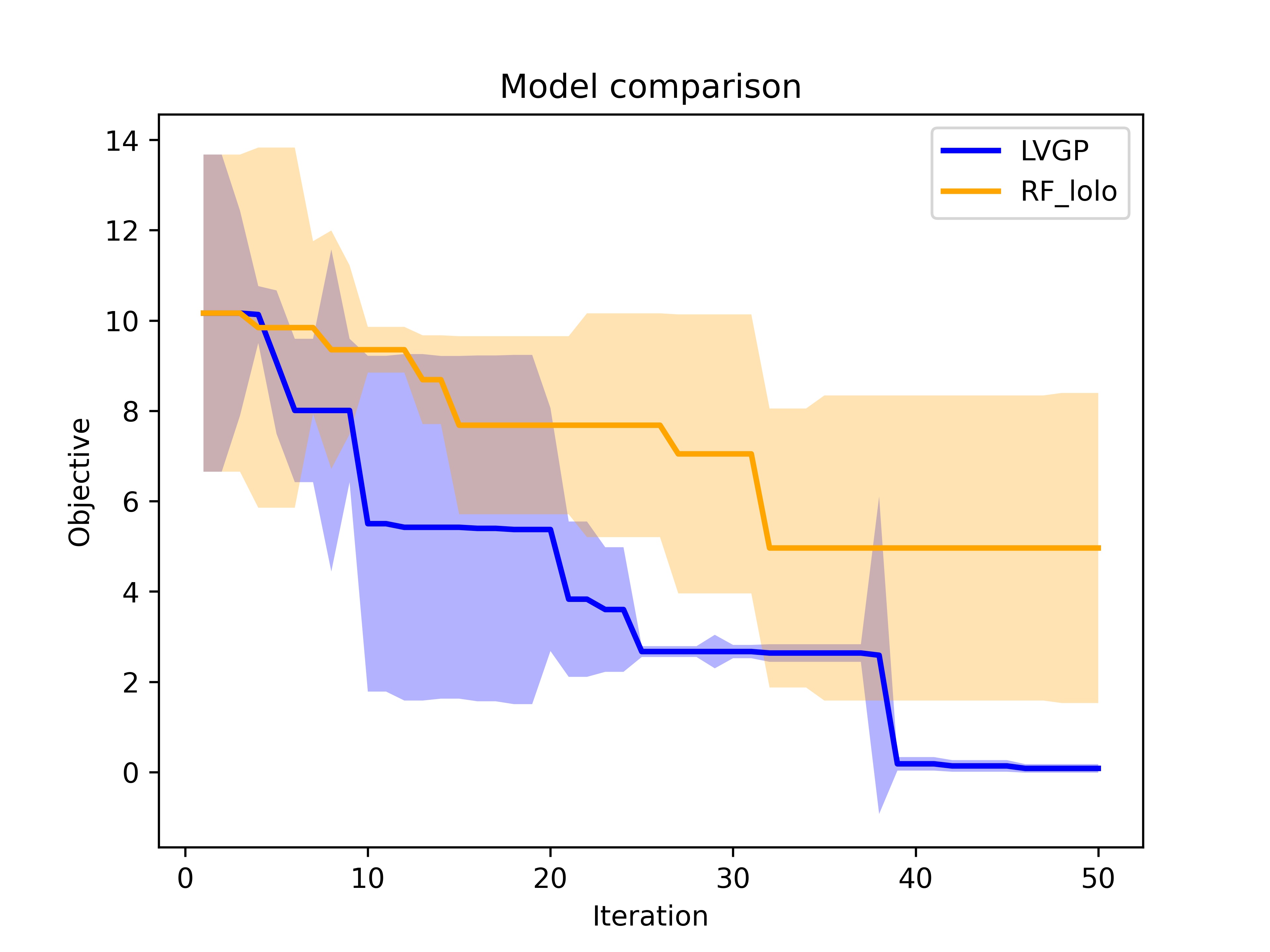}
    \caption{Optimization history plot of the Ackley function with initial sample size 65.}
    \label{fig:ackley}
\end{figure}

\paragraph{The Ackley function}
\begin{equation}
    f(\mathbf{v})=-20 \exp \left(-b \sqrt{\frac{1}{d} \sum_{i=1}^{d} v_{i}^{2}}\right)-\exp \left(\frac{1}{d} \sum_{i=1}^{d} \cos \left(c v_{i}\right)\right)+20+e,
\end{equation}
where $e=2.71828...$; we set the dimensionality $d=3$, with $v_{1,2} = x_{1,2} \in [-32.768, 32.768]$, and $v_3=t\in \{\pm 32, \pm 31, ..., 0\}$. Figure \ref{fig:ackley} is created from 5 runs.

\begin{figure}[H]
    \centering
    \includegraphics[width=0.5\textwidth]{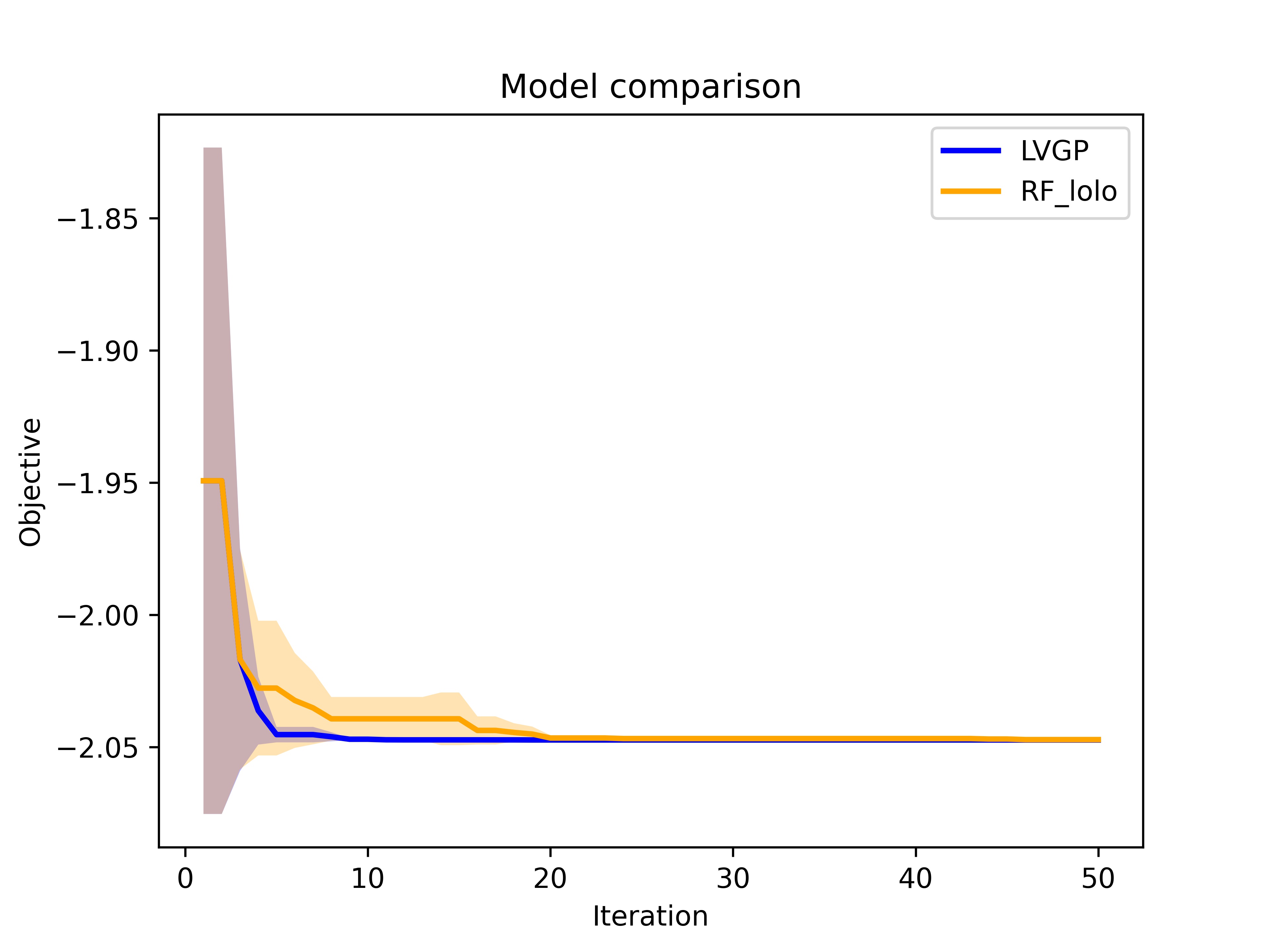}
    \caption{Optimization history plot of the Cross-in-Tray function with initial sample size 21.}
    \label{fig:crossintray}
\end{figure}

\paragraph{The Cross-in-Tray function}
\begin{equation}
    f(x,t)=-0.0001\left(\left|\sin \left(x\right) \sin \left(t\right) \exp \left(\left|100-\frac{\sqrt{x^{2}+t^{2}}}{\pi}\right|\right)\right|+1\right)^{0.1},
\end{equation}
where $x\in [-10, 10]$, and $t \in \{\pm 10, \pm 9, ..., 0\}$. Figure \ref{fig:crossintray} is created from 10 runs.

\begin{figure}[H]
    \centering
    \includegraphics[width=0.5\textwidth]{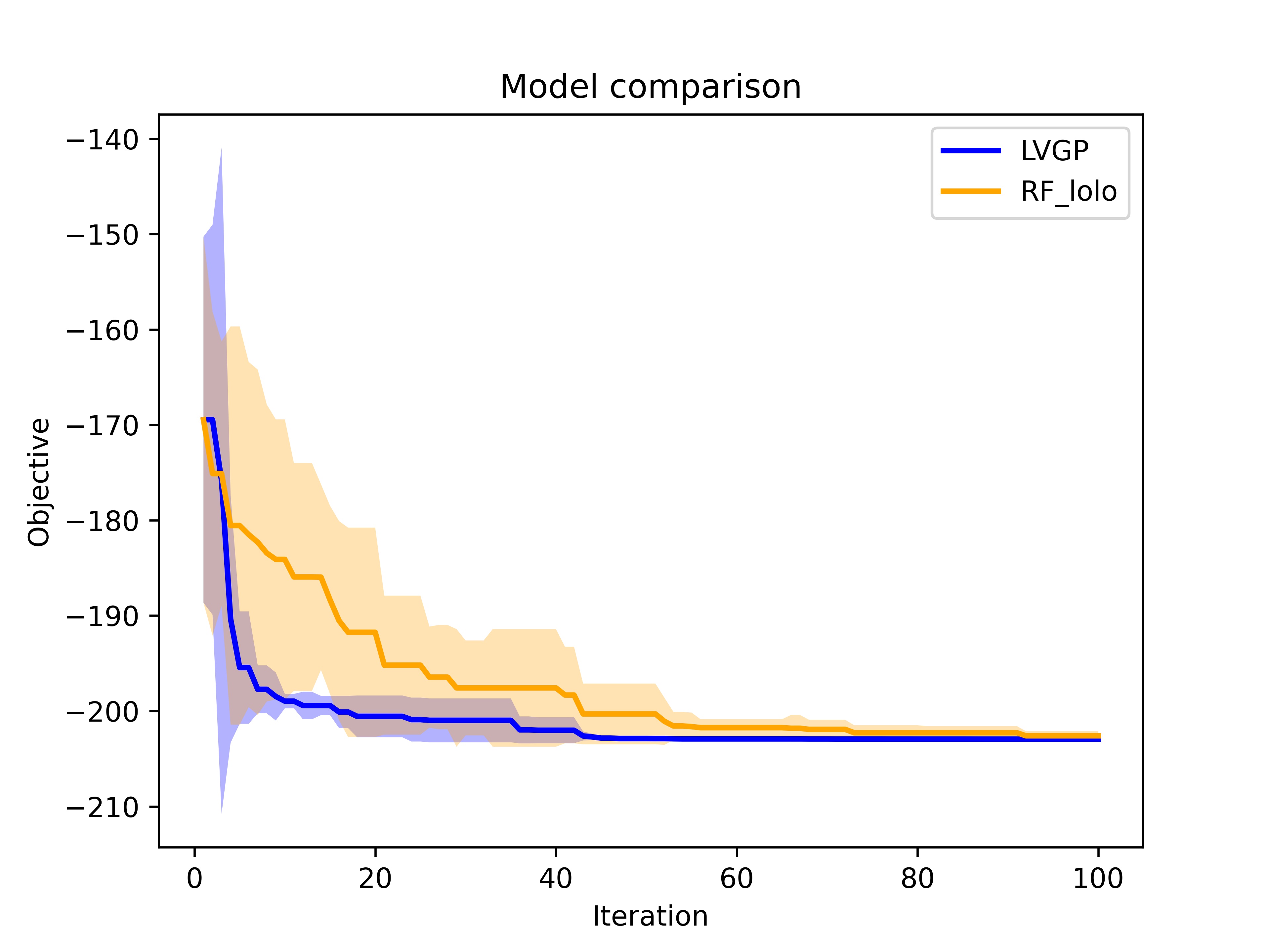}
    \caption{Optimization history plot of the Shubert function with initial sample size 21.}
    \label{fig:shubert}
\end{figure}

\paragraph{The Shubert function}
\begin{equation}
    f(x,t)=\left(\sum_{i=1}^{5} i \cos \left((i+1) x+i\right)\right)\left(\sum_{i=1}^{5} i \cos \left((i+1) t+i\right)\right),
\end{equation}
where $x\in [-10, 10]$, and $t \in \{\pm 10, \pm 9, ..., 0\}$. Figure \ref{fig:shubert} is created from 10 runs.

\paragraph{Sampling path}
Figures that show the sampling sequences by LVGP-BO and Lolo-BO in optimizing the Branin function are provided in a separate compressed file.

\section{Materials Property Test Cases}
\begin{figure}[H]
  \centering
  \begin{subfigure}[b]{0.48\textwidth}
    \centering
    \includegraphics[width=\textwidth]{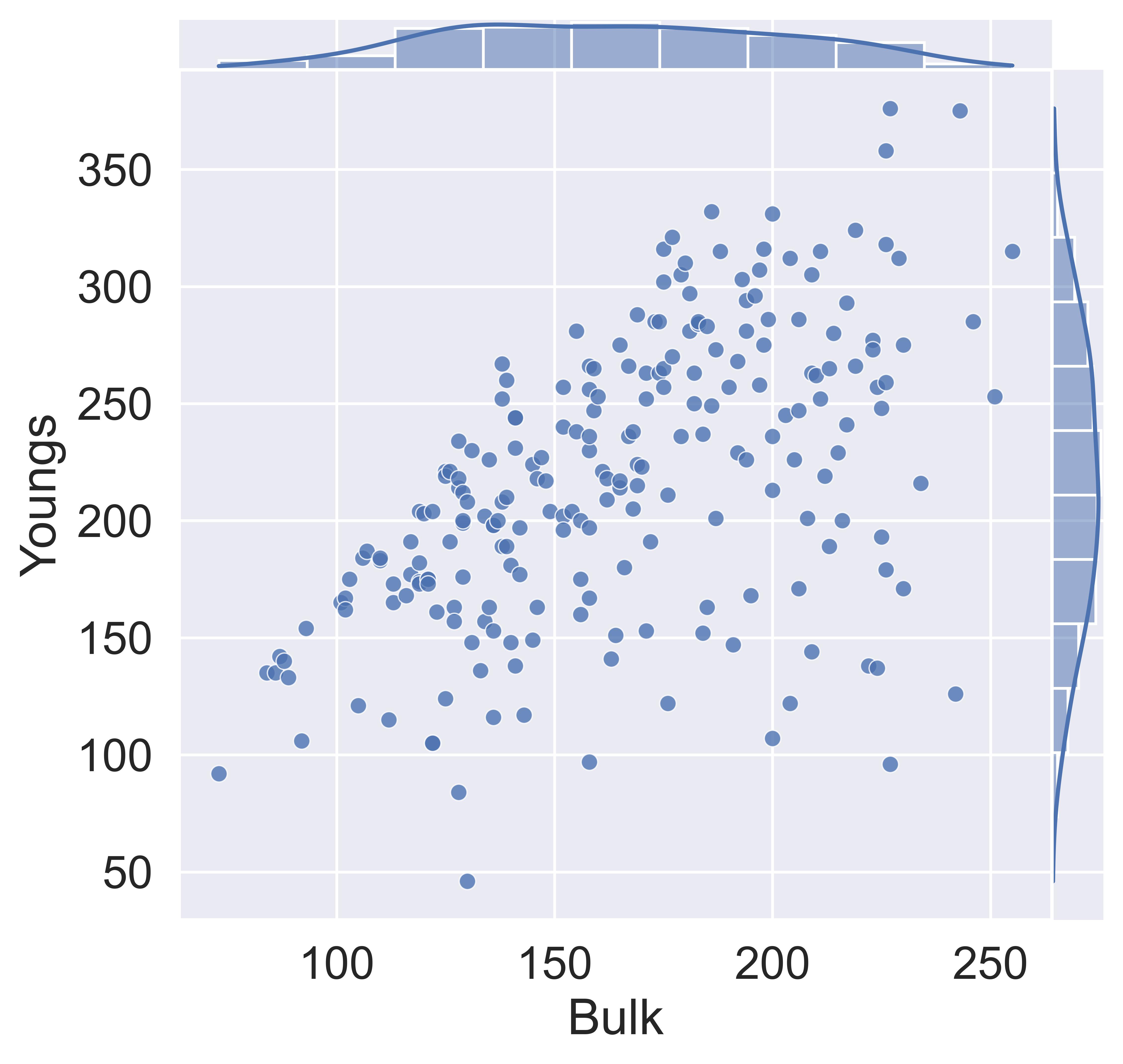}
  \end{subfigure}
  \begin{subfigure}[b]{0.48\textwidth}
    \centering
    \includegraphics[width=\textwidth]{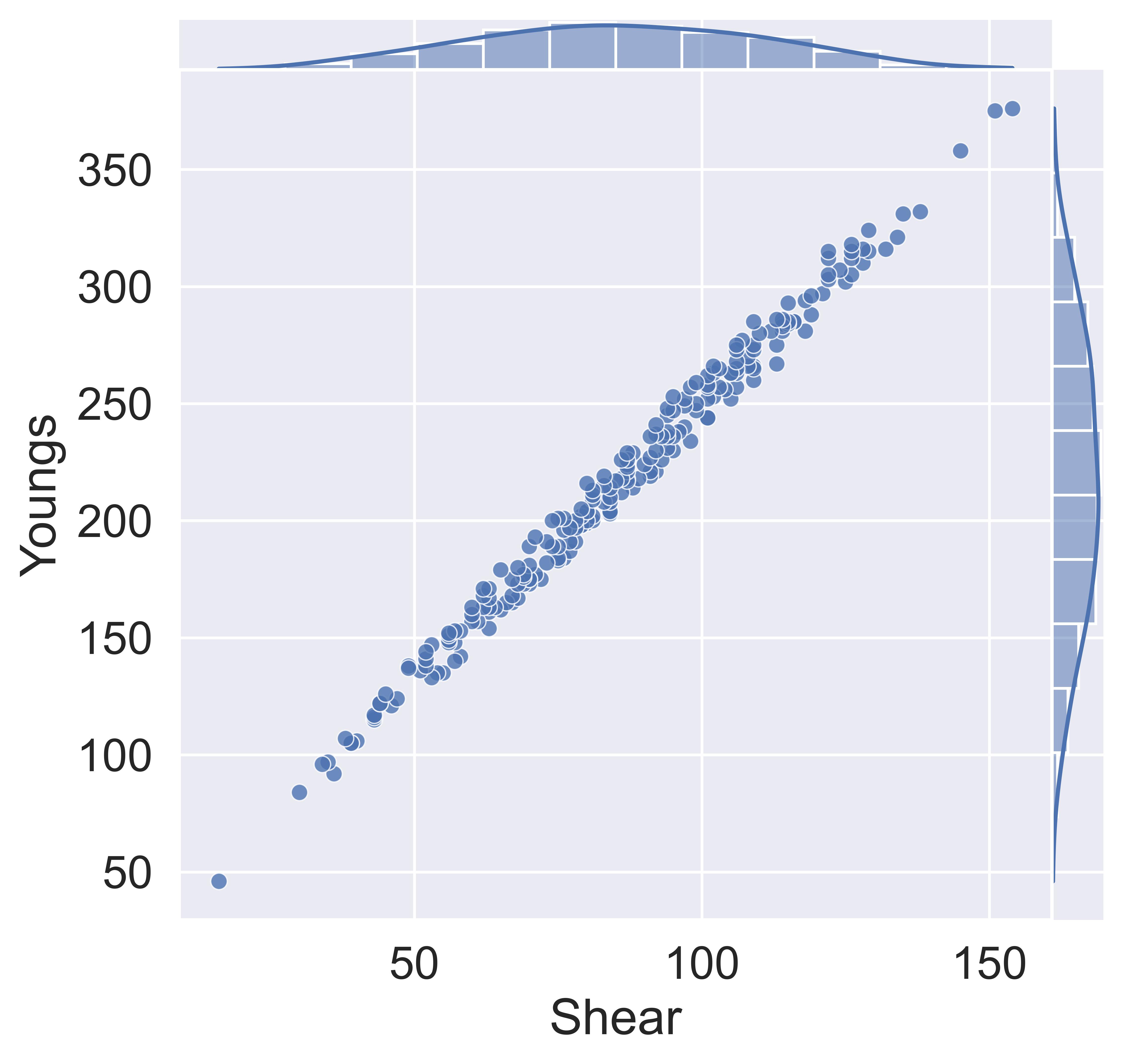}
  \end{subfigure}
  \caption{Scatter plots of Young's--bulk modulus ($E$--$B$) and Young's--shear modulus($E$--$G$).}
  \label{fig:moduli_plots}
\end{figure}

Figure \ref{fig:moduli_plots} showing the relations between Young's, shear, and bulk modulus of $\rm M_2AX$ compounds, indicate that Young's and shear moduli are highly correlated, while bulk modulus is not with the others.

\begin{figure}[H]
  \centering
  \begin{subfigure}[b]{0.58\textwidth}
    \centering
    \includegraphics[width=\textwidth]{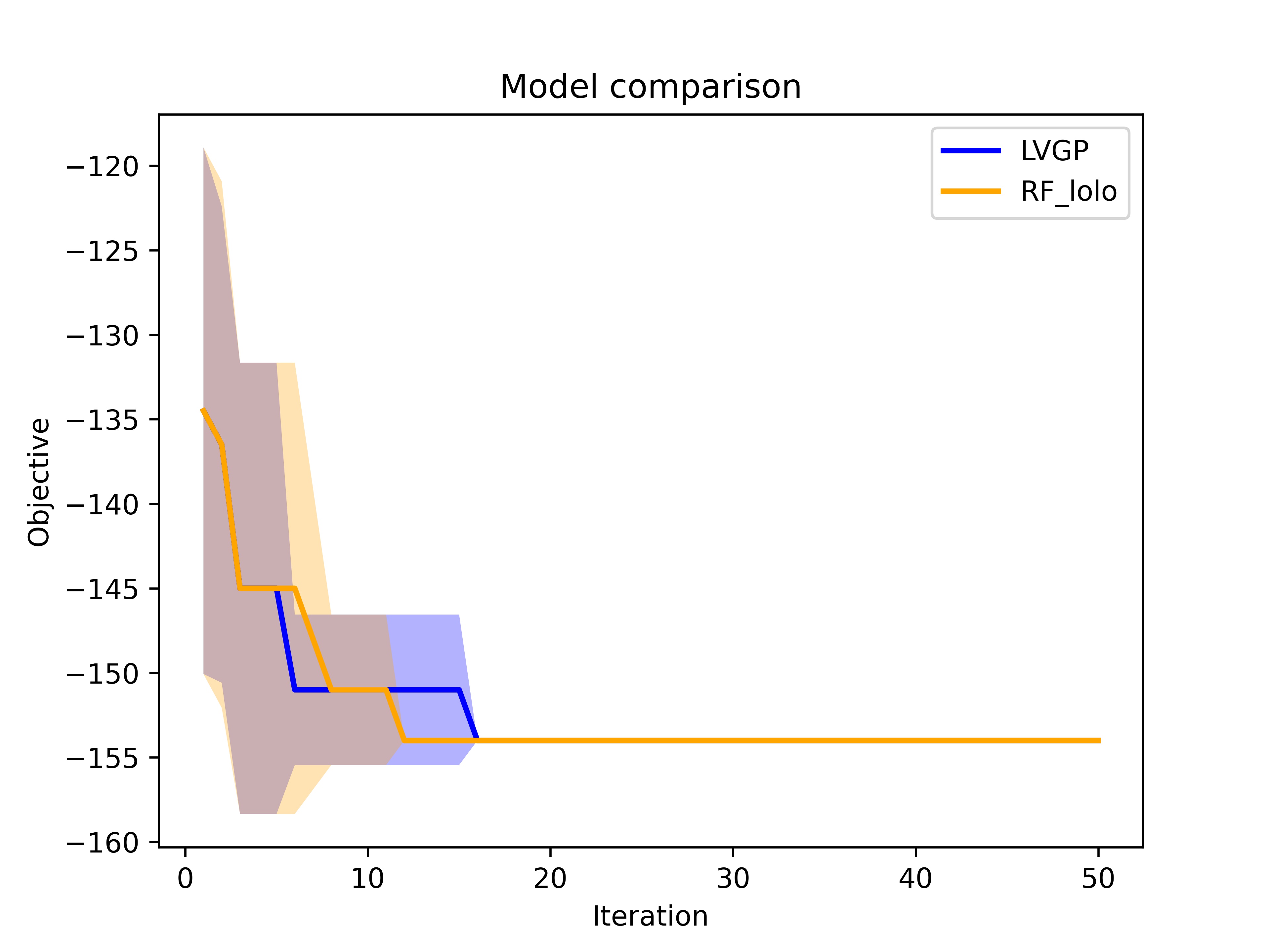}
  \end{subfigure}
  \begin{subfigure}[b]{0.4\textwidth}
    \centering
    \includegraphics[width=\textwidth]{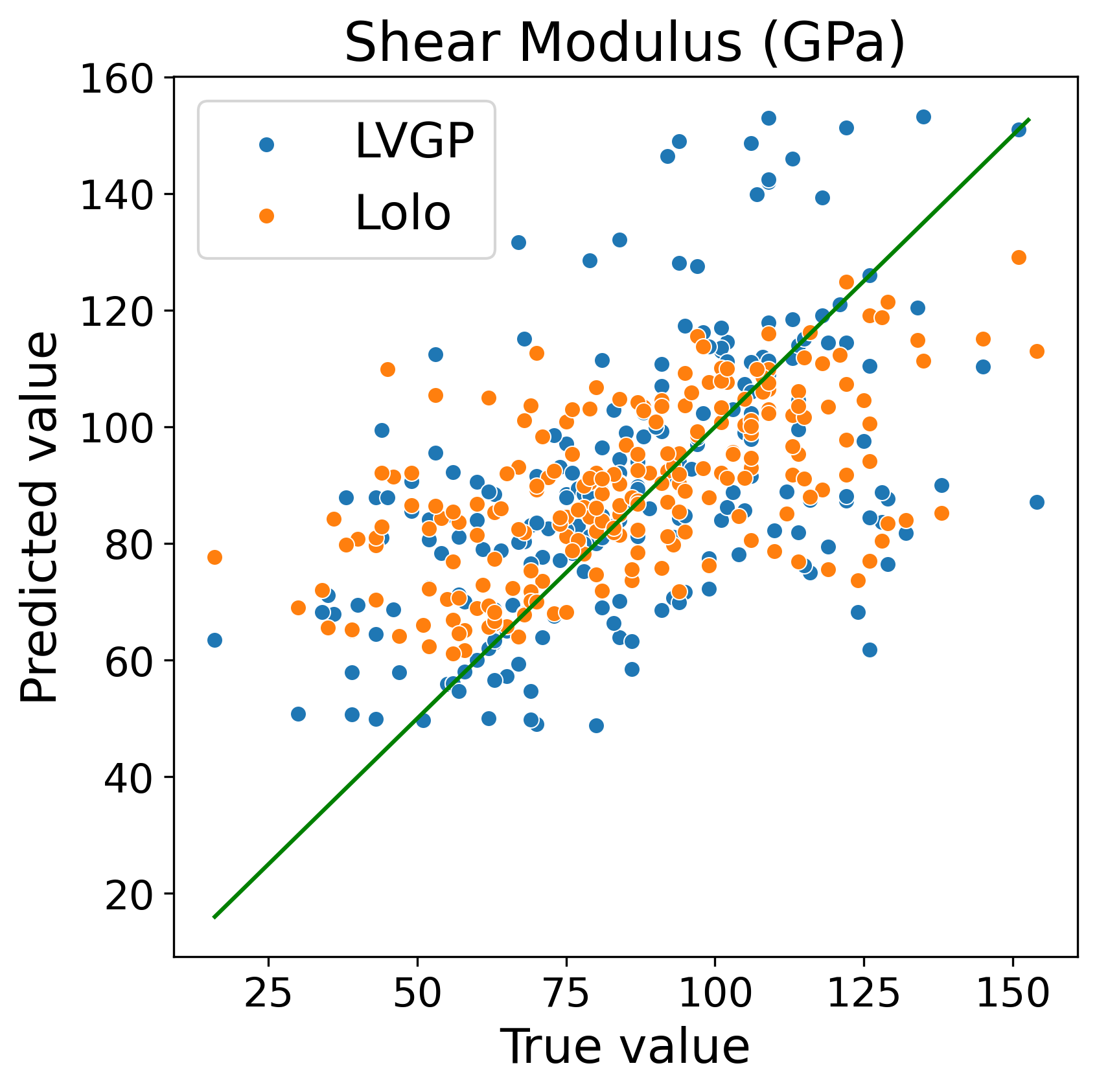}
  \end{subfigure}
  \caption{Optimization history plot (left) and regression plot (right) of $\rm M_2AX$ shear modulus, with 30 initial samples.}
  \label{fig:max_shear_30}
\end{figure}

Figure \ref{fig:max_shear_30} shows the BO and ML fitting results for the shear modulus of $\rm M_2AX$ compounds.
